\documentclass[lettersize,journal]{IEEEtran}
\usepackage{amsmath,amsfonts,amssymb}
\usepackage{algorithmic}
\usepackage{algorithm}
\usepackage{array}
\usepackage{acronym}
\usepackage{multirow}
\usepackage{textcomp}
\usepackage{stfloats}
\usepackage{url}
\usepackage{verbatim}
\usepackage{bbm}
\usepackage{xcolor}
\usepackage{color}
\usepackage{tabularray}
\usepackage{graphicx}
\usepackage{cite}
\usepackage{orcidlink}
\usepackage{cleveref}
\usepackage{subfigure}
\usepackage{booktabs}  
\usepackage{tabularx}  
\usepackage{siunitx}   
\usepackage{arydshln} 

\newcommand{\cy}[1]{\textcolor{black}{#1}}

\definecolor{Alto}{rgb}{0.93,0.93,0.93}

\usepackage[switch]{lineno}

\begin{document}

\title{Adaptive Gradient Calibration for Single-Positive Multi-Label Learning in Remote Sensing Image Scene Classification}

\author{Chenying~Liu~\orcidlink{0000-0001-9172-3586},~\IEEEmembership{Graduate~Student~Member,~IEEE},
        Gianmarco~Perantoni~\orcidlink{0000-0002-1146-1035},~\IEEEmembership{Graduate~Student~Member,~IEEE},
        Lorenzo~Bruzzone~\orcidlink{0000-0002-6036-459X},~\IEEEmembership{Fellow,~IEEE},
        and~Xiao~Xiang~Zhu~\orcidlink{0000-0001-5530-3613},~\IEEEmembership{Fellow,~IEEE}
\thanks{Chenying Liu and Gianmarco Perantoni contributed equally to this work. \textit{(Corresponding authors:
Lorenzo Bruzzone; Xiao Xiang Zhu.)}}
\thanks{Chenying Liu and Xiao Xiang Zhu are with the Chair of Data Science in
Earth Observation, Technical University of Munich (TUM), and Munich Center for Machine Learning (MCML), 80333 Munich, Germany (e-mail: chenying.liu@tum.de; xiaoxiang.zhu@tum.de).}
\thanks{Gianmarco Perantoni and Lorenzo Bruzzone are with the Department of Information Engineering and Computer Science, University of Trento, 38123 Trento, Italy (e-mail: gianmarco.perantoni@unitn.it; lorenzo.bruzzone@unitn.it).}
\thanks{This work has been submitted to the IEEE for possible publication. Copyright may be transferred without notice, after which this version may no longer be accessible.}
}


\markboth{Under review for publication in IEEE Transactions on Geoscience and Remote Sensing}%
{Shell \MakeLowercase{\textit{et al.}}: A Sample Article Using IEEEtran.cls for IEEE Journals}


\maketitle

\begin{abstract}
Multi-label classification (MLC) offers a more comprehensive semantic understanding of Remote Sensing (RS) imagery compared to traditional single-label classification (SLC). However, obtaining complete annotations for MLC is particularly challenging due to the complexity and high cost of the labeling process. As a practical alternative, single-positive multi-label learning (SPML) has emerged, where each image is annotated with only one relevant label, and the model is expected to recover the full set of
labels. While scalable, SPML introduces significant supervision ambiguity, demanding specialized solutions for model training. Although various SPML methods have been proposed in the computer vision domain, research in the RS context remains limited. To bridge this gap, we propose Adaptive Gradient Calibration (AdaGC), a novel and generalizable SPML framework tailored to RS imagery. AdaGC adopts a gradient calibration (GC) mechanism
\cy{with a dual exponential moving average (EMA) module} for robust pseudo-label generation.
\cy{We introduce a theoretically grounded, training-dynamics-based indicator to adaptively trigger GC, which ensures GC's effectiveness by preventing it from being affected by model underfitting or overfitting to label noise.} 
Extensive experiments on two benchmark RS datasets under two distinct label noise types demonstrate that AdaGC achieves state-of-the-art (SOTA) performance while maintaining strong robustness across diverse settings.
\cy{The codes and data will be released at \url{https://github.com/rslab-unitrento/AdaGC}.}
\end{abstract}



\newacro{AdaGC}{Adaptive Gradient Calibration}
\newacro{GC}{Gradient Calibration}
\newacro{AN}{Assume Negative}
\newacro{AN-LS}{\ac{AN} with label smoothing} 
\newacro{AP}{average precision}
\newacro{BCE}{binary cross-entropy}
\newacro{mAP}{mean average precision}
\newacro{CAM}{class activation map}
\newacro{CCE}{categorical cross-entropy}
\newacro{CLC}{CORINE Land Cover} 
\newacro{CNN}{Convolutional Neural Network}
\newacro{ConvLSTM}{Convolutional Long Short-Term Memory}
\newacro{CORINE}{COoRdination of INformation on the Environment} 
\newacro{CV}{computer vision}
\newacro{DL}{deep learning}
\newacro{ELR}{Early Learning Regularization}
\newacro{EMA}{exponential moving average}
\newacro{EO}{Earth observation}
\newacro{EPR}{Expected Positives Regularization}
\newacro{mF1}{mean F1 score}
\newacro{FCL}{fully connected layer}
\newacro{FM}{foundation model}
\newacro{GR Loss}{Generalized Robust Loss}
\newacro{GT}{ground-truth}
\newacro{mIoU}{mean intersection over union}
\newacro{IU}{Ignoring Unobserved}
\newacro{IUN}{\ac{IU} with true Negatives}
\newacro{LAGC}{Label-Aware Global Consistency}
\newacro{LC}{land-cover}
\newacro{LL}{large loss}
\newacro{LSTM}{Long Short-Term Memory}
\newacro{LULC}{land-use and land-cover}
\newacro{LU}{land-use}
\newacro{MIME}{Mutual label enhancement for sIngle-positive Multi-label lEarning}
\newacro{MGRS}{Military Grid Reference System} 
\newacro{ML}{maximum likelihood}
\newacro{MCC}{multi-class classification}
\newacro{MLC}{multi-label classification}
\newacro{OA}{overall accuracy}
\newacro{PUL}{positive-unlabeled learning}
\newacro{RF}{Random Forest}
\newacro{RNG}{random number generator}
\newacro{RNN}{Recurrent Neural Network}
\newacro{RS}{remote sensing}
\newacro{reBEN}{refined BigEarthNet} 
\newacro{ResNet}{Residual neural Network}
\newacro{ROLE}{Regularized Online Label Estimation}
\newacro{SAR}{synthetic aperture radar}
\newacro{SPML}{single-positive multi-label learning}
\newacro{SLC}{single-label classification}
\newacro{S1}{Sentinel-1} 
\newacro{S2}{Sentinel-2} 
\newacro{SVM}{Support Vector Machine}
\newacro{SOTA}{state-of-the-art}
\newacro{TPE}{Tree-structured Parzen Estimator}
\newacro{TS}{time series}
\newacro{ViT}{Vision Transformer}
\newacro{VHR}{very-high resolution}
\newacro{VIB}{Variational Information Bottleneck}
\newacro{WAN}{Weak \ac{AN}}
\newacro{WSML}{weakly supervised multi-label learning}

\section{Introduction}\label{sec:intro}
\begin{figure}[tp]
    \centering
    \scriptsize
    \begin{tabular}{c|cc}
        \includegraphics[width=0.28\linewidth]{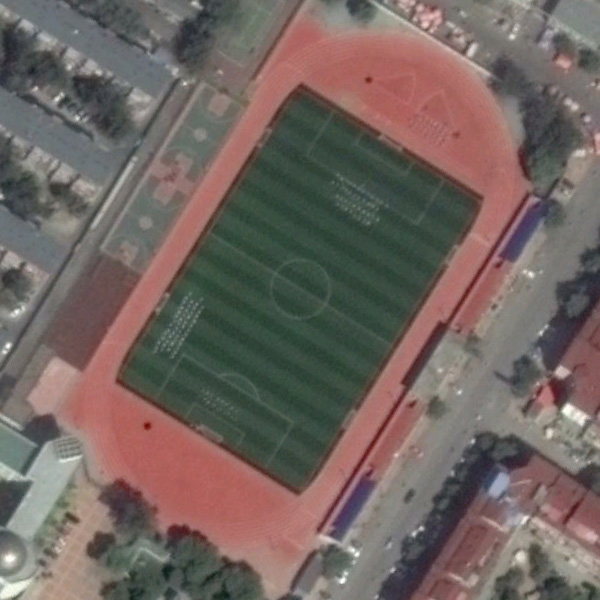} & 
        \includegraphics[width=0.28\linewidth]{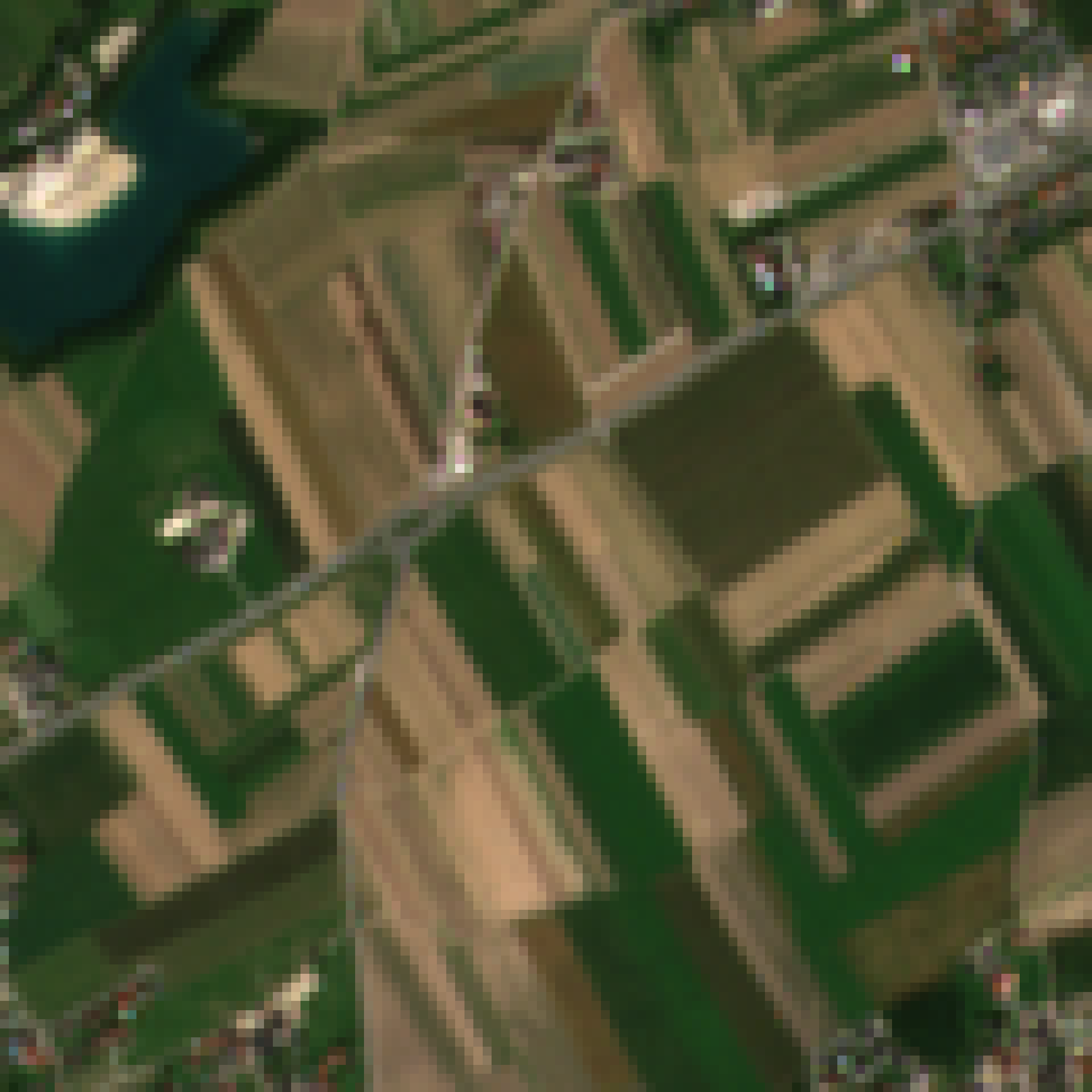} &
        \includegraphics[width=0.28\linewidth]{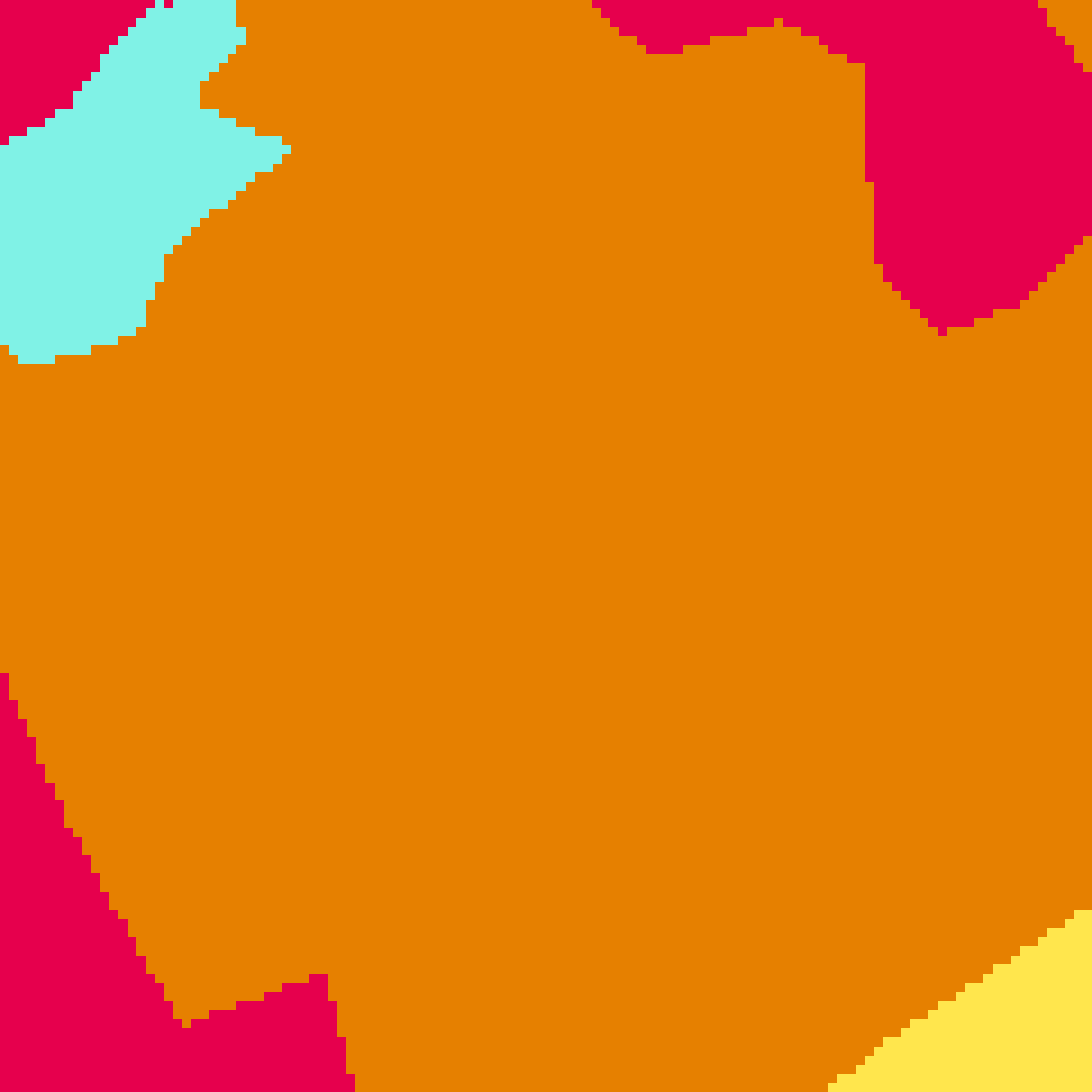} \\
        \parbox{0.3\linewidth}{
            \centering
            (a) \textbf{Single-label}: court \\
            \textbf{Multi labels}: buildings,\\
            cars, pavement, court, \\
            grass, trees
        } &
        \multicolumn{2}{c}{
        \parbox{0.6\linewidth}{
            \centering
            (b) \textbf{Single-label}: arable land (dominant) \\
            \textbf{Multi-labels}: \textcolor[HTML]{E68000}{arable land},\\
            \textcolor[HTML]{E6004D}{urban fabric}, \textcolor[HTML]{80F2E6}{inland waters}, \\
            \textcolor[HTML]{FFE64D}{complex cultivation patterns}
        }}
    \end{tabular}
    \caption{Single- and multi-label annotation examples from (a) AID-multilabel \cite{hua_relation_2020} and (b) refined BigEarthNet \cite{clasen_reben_2025_arxivonly} datasets, where the corresponding CLC mask is presented for reference. Compared to the single-class labels, the multi-label annotations can more comprehensively describe the scene.}
    \label{fig:intro:mlc-example}
\end{figure}

\IEEEPARstart{W}{ith} the rapid development of satellite missions, vast amounts of \ac{RS} imagery have become increasingly accessible, serving as a critical data source for numerous applications \cite{zhao_artificial_2024, zhu_foundations_2024}. Image classification \cite{mehmood_remote_2022, dimitrovski_current_2023, he_recent_2018}, a fundamental task to connect \ac{RS} imagery with downstream applications, still largely adopts the \ac{SLC} paradigm. However, assigning a single label to each image oversimplifies the complexity of real-world \ac{RS} scenes \cite{Hua_2019_JPRS, Sumbul_2020_ACCESS}. In practice, \ac{RS} images often contain diverse land cover types, objects, or functional zones. As illustrated in Fig.~\ref{fig:intro:mlc-example}, both VHR and lower-resolution imagery commonly capture heterogeneous content, making \ac{MLC} a more informative representation than \ac{SLC}.

A key challenge for \ac{MLC} is the difficulty and cost of acquiring complete, accurate annotations \cite{Burgert_2022_TGRS}, especially in the \ac{DL} era where large-scale labeled datasets are crucial \cite{zhu_deep_2017}. A practical workaround is to annotate only one relevant label per image, leading to the \ac{SPML} setting \cite{Cole_2021_CVPR}. In this setting, the model is weakly supervised with one positive label per image, yet it is expected to recover the complete set of relevant labels. The supervision in \ac{SPML} is characterized by missing labels. In practice, even expert annotators often miss relevant labels, making the false negative issue pervasive in \ac{MLC}. \ac{SPML} can thus be viewed as an extreme case of this problem. 

Although \ac{SPML} is promising in scalability and annotation efficiency, it introduces significant ambiguity. 
A common baseline approach is the \ac{AN} strategy \cite{Cole_2021_CVPR}, which treats all unobserved labels as negatives, unavoidably introducing bias into learning. In the \ac{CV} domain, several strategies have been proposed to address its limitations, such as noise-robust regularization terms, online label-correction/pseudo-labeling methods, and interpretability-driven frameworks \cite{Cole_2021_CVPR, Zhang_2018_ICLR, Kim_2022_CVPR, Xie_2022_NIPS, Chen_2024_IJCAI, Kim_2023_CVPR, Liu_2023_ICML}. Despite the advancements in \ac{CV}, the \ac{SPML} problem remains underexplored in \ac{RS}. \ac{SPML} is particularly valuable for \ac{RS} image classification, where annotation typically requires greater domain expertise and incurs higher costs.
These challenges make the application of \ac{SPML} in \ac{RS} more necessary and also more difficult.

When directly applying existing \ac{SPML} methods to \ac{RS} data, several challenges arise. One major issue is the reliability of pseudo-labels. \ac{VHR} \ac{RS} images often include small, sparsely distributed targets, while medium-to-low-resolution imagery tends to capture broader regional semantics. Both cases can easily result in low inter-class separability and high intra-class variability, hindering the effectiveness of simple pseudo-labeling strategies such as confidence thresholding. Furthermore, the distinct characteristics of \ac{MLC} versus \ac{MCC} suggest that commonly used learning-from-noisy-labels strategies, which are often designed for \ac{MCC}, should be carefully tailored to \ac{SPML}.
\cy{Moreover, most existing pseudo-label-based \ac{SPML} methods depend on fixed warm-up schedules, which often result in suboptimal performance due to model underfitting or overfitting to label noise.}

In this work, we propose a novel and generalizable framework for \ac{SPML} in \ac{RS}, termed \ac{AdaGC}, to address the aforementioned challenges. The framework consists of three key components. 
\cy{First, we propose a theoretically grounded Gradient Calibration (GC) mechanism designed specifically for \ac{SPML}. GC aims to alleviate the model's tendency to overfit false negatives by correcting gradients using pseudo-label guidance. Unlike Early Learning Regularization (ELR) \cite{liu_early-learning_2020-2}, which is formulated for multi-class classification under the simplex constraint, GC is rigorously extended to the SPML setting without such a constraint. Second, to guarantee pseudo-label quality, we introduce a dual \ac{EMA} module for robust pseudo-label estimation, which leverages the consensus between the temporally smoothed predictions of the student model and the teacher model's outputs. The former applies \ac{EMA} to predictions, and the latter to model weights. }
\cy{Then, to ensure the effectiveness of GC, we conduct an in-depth analysis of learning dynamics in \ac{SPML} and propose a theoretically grounded indicator for adaptive activation of \ac{GC} after an initial warm-up stage. This mechanism is central to our design, which is capable of preventing interference from underfitting or noise-driven overfitting to stabilize optimization.}
\cy{We also incorporate the Mixup data augmentation technique \cite{Zhang_2018_ICLR} within \ac{GC} to further improve model's generalizability.} 
Extensive experiments on two benchmark \ac{RS} multi-label datasets demonstrate the effectiveness and robustness of \ac{AdaGC}.

The main contributions of this paper are summarized as follows:
\begin{itemize}
    \item \cy{We propose a two-stage \ac{RS} \ac{SPML} framework, which incorporates an early learning detection strategy to guide the activation of \ac{GC} to combat underfitting and overfitting to label noise. The theoretical analyses are also provided to support the effectiveness of the framework design.}
    \item We introduce a dual-EMA pseudo-label generation strategy that enhances label completeness and reliability by leveraging temporal prediction fusion.
    \item \cy{We conduct extensive experiments on two benchmark \ac{RS} datasets covering both high- and low-resolution imagery. For low-resolution data, we employ \textit{Random} and \textit{Dominant} SPML noise simulations to better reflect real-world annotation behavior. For high-resolution imagery, where identifying a dominant class is inherently difficult, we rely on \textit{Random} simulation supplemented with additional \textit{Manual} single-positive annotations. We provide a comprehensive benchmarking of existing SPML methods, offering valuable insights for future research in this domain.}
\end{itemize}

The remainder of this paper is organized as follows. Section~\ref{sec:relatedworks} reviews existing works on \ac{MLC}, with a focus on noisy labels and \ac{SPML}. Section~\ref{sec:method} details the proposed method along with the theoretical analyses. The empirical evaluation is presented in Section~\ref{sec:experiments}, with the conclusion in Section~\ref{sec:con}.

\section{Related Works}\label{sec:relatedworks}
This section reviews advancements in \ac{MLC}, with emphasis on \ac{RS} applications, challenges from noisy labels, and recent \ac{SPML} methods developed in \ac{CV}.

\subsection{Remote Sensing Image Multi-label Classification}

Early \ac{DL} research in MLC for RS imagery primarily focused on high-resolution data and employed \acp{CNN} for feature extraction, followed by classifiers such as Radial Basis Function Neural Networks \cite{Zeggada_2017_GRSL} or Structured Support Vector Machines \cite{Koda_2018_TGRS}. These approaches often relied on conventional transfer learning, using \acp{CNN} pre-trained on generic \ac{CV} datasets (\textit{e.g.}, ImageNet \cite{Russakovsky_2015_IJCV}) as fixed feature extractors. However, the significant differences between natural \ac{CV} scenes and complex \ac{RS} imagery can lead to suboptimal classification performance, limiting the effectiveness of such direct transfer.

To mitigate the limitations of pre-trained models, alternative strategies have been proposed. For instance, Stivaktakis \textit{et al.} \cite{Stivaktakis_2019_GRSL} introduced a data augmentation scheme to facilitate end-to-end training of shallow \acp{CNN}, along with replacing the softmax layer with a sigmoid function to accommodate multi-label outputs. While this direct adaptation remains a common practice in multi-label scenarios, it may result in the imprecise identification of multiple classes. 
More sophisticated approaches integrate sequential neural network architectures with \acp{CNN} to enhance multi-label scene classification. Hua \textit{et al.} \cite{Hua_2019_JPRS} proposed a class-wise attention-based \ac{RNN} to explicitly model the co-occurrence relationships among multiple classes, thus generating class predictions sequentially. Similarly, an attention-aware label relational reasoning network was introduced in \cite{hua_relation_2020} to localize discriminative regions and characterize inter-label relationships based on feature maps. Alshehri \textit{et al.} \cite{Alshehri_2019_ACCESS} presented an encoder-decoder neural network architecture, where the encoder includes a squeeze-and-excitation layer to model channel-wise dependencies, and the RNN-based decoder serves as an adaptive spatial attention module. These attention-driven strategies have demonstrated considerable success in identifying informative areas through attention maps derived from convolutional features.

\subsection{Multi-label Classification with Noisy Labels}

A major hurdle in \ac{MLC} in \ac{RS} is the difficulty in obtaining exhaustive multi-label annotations, which often results in noisy labels. This noise can be categorized as either missing labels (subtractive noise), where a class present in the considered image is not annotated, or as wrong labels (additive noise), where an absent class is incorrectly annotated \cite{Burgert_2022_TGRS}. Such multi-label noise, which can be introduced through zero-cost annotation (\textit{e.g.}, using thematic products), can severely impair model training. Consequently, developing methods robust to multi-label noise in \ac{RS} is an active research area, usually categorized as \ac{WSML} \cite{Xie_2023_TPAMI}.

Methods for handling noisy labels in multi-label settings often draw inspiration from related fields, such as \ac{PUL} \cite{duPlessis_2014_NIPS,Gang_2016_NIPS,bekker_2020_ML} and noisy multi-class learning \cite{perantonirobust}. Common strategies involve developing robust loss functions that are less sensitive to label noise or class imbalance \cite{li2020robust}. Another approach is pseudo-labeling, where likely labels for unobserved or unknown annotations are inferred to provide additional training signals \cite{Zhang_2020_RS}. 
The “memorization effect,” where models tend to learn clean labels early but later start to memorize noisy ones, has been widely observed in noisy \ac{MCC}. \ac{ELR} \cite{Liu_2020_NIPS} was developed to specifically counteract this, by adjusting the influence of gradients from potentially noisy samples, particularly in the early stages of training. Burgert \textit{et al.} \cite{Burgert_2022_TGRS} later extended it to MLC in a straightforward manner and empirically demonstrated its effectiveness.
Additional strategies include sample selection, which aims to identify and potentially exclude noisy labels \cite{Paris_2021_TGRS}, label correction/refurbishment, which attempts to correct noisy labels (\textit{e.g.,} using collaborative learning frameworks \cite{Aksoy_2021_ICIP,Aksoy_2024_TNNLS}), \cy{and noise-robust regularization, such as using word embedding similarities to regularize model training \cite{hua_learning_2020}.}

Other methods, such as those in the \ac{LL} \cite{Kim_2022_CVPR} family, including LL-R, LL-Ct, and LL-Cp, are developed to recognize and handle mislabeled samples with large losses, based on the observation that noisy labels often lead to high loss values.
Specifically, LL-R rejects samples with large losses, LL-Ct applies temporary corrections, and LL-Cp employs permanent corrections. BoostLU \cite{Kim_2023_CVPR} complements \ac{LL} by enhancing predicted positive regions via a function applied to \acp{CAM}, further improving the robustness.

In \ac{WSML} for \ac{RS}, subtractive noise, \textit{i.e.}, missing true labels, is a prevalent issue. This often occurs because annotators, especially when facing complex imagery or large datasets, opt to label only the dominant class(es) rather than exhaustively identify all present categories.
This poses a significant challenge because an unannotated class doesn't necessarily mean its absence, leading to the extreme case of subtractive noise known as \ac{SPML}. This scenario is both challenging and increasingly common in practical \ac{RS} applications, despite being underexplored in the \ac{RS} literature.

\subsection{Single Positive Multi-Label Learning}
\ac{SPML} was formally defined by Cole \textit{et al.} \cite{Cole_2021_CVPR} as the scenario where each training image is annotated with only one positive label, while all other labels remain unobserved.
A naive approach, such as \ac{IU} labels, is insufficient in \ac{SPML} and typically results in the model collapsing to predicting all labels as positive because of the complete lack of negative constraints.
Therefore, the most common baseline in \ac{SPML} is the \ac{AN} loss, which treats all unobserved labels as negative during training. While straightforward, this assumption inherently introduces false negative labels, which can negatively impact model performance and generalization. 
To mitigate these challenges, Cole \textit{et al.} \cite{Cole_2021_CVPR} proposed several modified baseline methods for \ac{MLC}. For instance, \ac{AN-LS} applies label smoothing to reduce the penalty associated with potentially incorrect negative labels. \Ac{WAN} introduces a weighting mechanism to down-weight the contribution of assumed negative labels in the loss computation. 
Other methods focus on regularizing model predictions or inferring missing labels. \Ac{EPR} is a regularization technique for \ac{IU} that constrains the expected number of positive labels per image, helping in preventing the trivial solution of predicting all labels as positive. Building upon this, \ac{ROLE} \cite{Cole_2021_CVPR} treats unannotated labels as learnable parameters, estimated online during training and regularized with \ac{EPR}.

Pseudo-labeling is another widely adopted strategy in \ac{SPML}.
In this case, a \ac{LAGC} \cite{Xie_2022_NIPS} regularization term is introduced to recover pseudo-labels, leveraging the data manifold structure learned through contrastive learning and data augmentation. \ac{GR Loss} \cite{Chen_2024_IJCAI} provides a unified approach to \ac{SPML} that incorporates soft pseudo-labels into a robust loss design to handle false negatives and class imbalance. \Ac{MIME} \cite{Liu_2023_ICML}, based on the \ac{VIB} concept \cite{Alemi_2017_ICLR}, is an iterative method that generates and refines pseudo-labels by maximizing the mutual information between model predictions and estimated pseudo-labels. 
Methods from \ac{PUL} are also relevant, which focus on the single binary classification counterpart of \ac{SPML}. The basic idea is to exploit an unbiased estimator of the empirical risk, which is only possible with the knowledge of the prior distribution of the positive class. For example, Dist-PU \cite{Zhao_2022_CVPR} exploits the positive class prior, along with Mixup \cite{Zhang_2018_ICLR}, to address negative prediction bias and encourage confident predictions. While designed for \ac{PUL}, it can be easily employed in \ac{SPML} by independently applying it to each class.

However, the above-mentioned methods are mainly developed for natural images, and their direct application to \ac{RS} data often falls short due to the distinct characteristics of \ac{RS} imagery. Moreover, while pseudo-labeling has shown effective, it typically requires a warm-up stage to ensure label quality. Manually determining this stage can be suboptimal, either due to insufficient learning or overfitting to noisy labels.

\section{Methodology}\label{sec:method}

\begin{figure*}
    \centering
    \includegraphics[width=.97\linewidth]{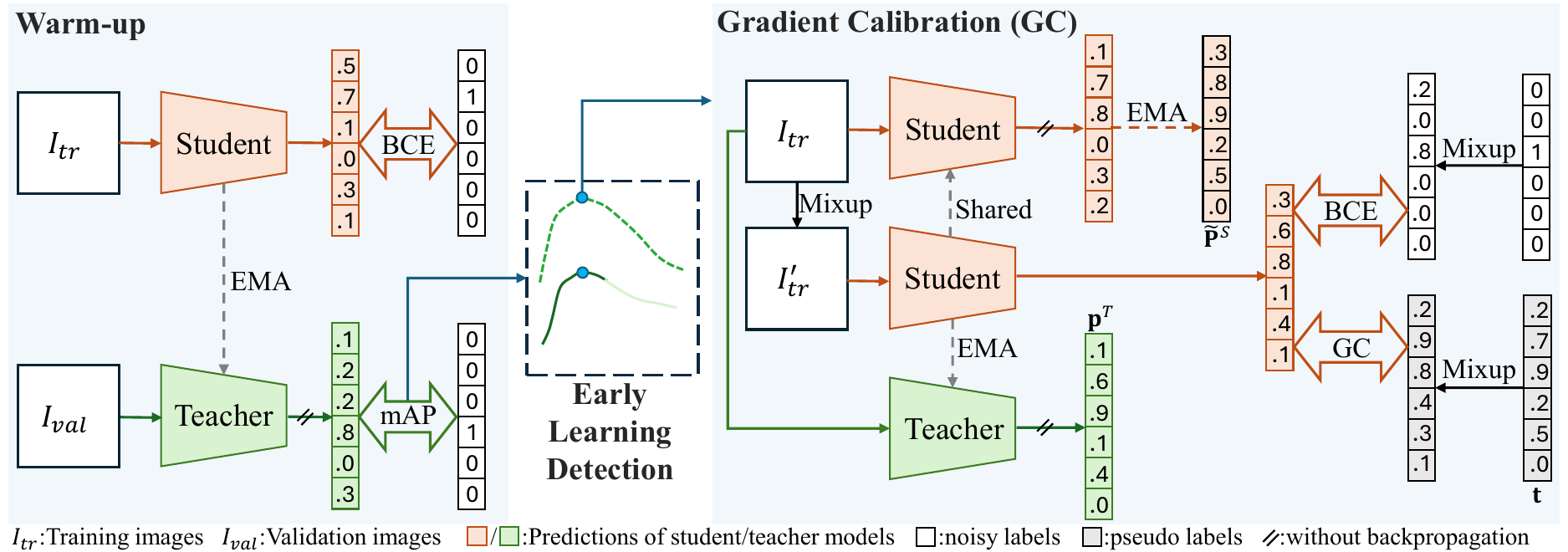}
    \caption{\cy{Flowchart of the proposed Adaptive Gradient Calibration (AdaGC) method for single-positive multi-label learning in remote sensing image classification. Pseudo-labels $\mathbf{t}$ are generated by combining the teacher model’s predictions $\mathbf{p}^T$ and the student model’s predictions $\tilde{\mathbf{p}}^S$ according to \eqref{eq:pseudo-label}. EMA is also applied to the student model’s predictions during the warm-up stage, which is omitted for simplification.}}
    \label{fig:meth:flowchart}
\end{figure*}

As illustrated in Fig. \ref{fig:meth:flowchart}, the proposed method adopts a student–teacher architecture, where the teacher model’s weights are updated via \ac{EMA}. The training pipeline consists of two stages. In the first stage, the student model is trained directly on noisy labels, during which the teacher model’s validation accuracy (with respect to noisy labels) is monitored as an early learning indicator to determine when to activate the second stage—\acf{GC}. Next, we first define the problem before presenting the technical details of AdaGC.

\subsection{Problem Definition} \label{sec:meth:problem}

In \ac{SPML}, each image $\mathbf{x}_i \in \mathbb{R}^{h\times w\times d}$ is associated with a true label vector $\mathbf{y}_i^* \in \{0,1\}^C$, where multiple labels may be relevant. However, during training, only a single positive label is observed per image. The observed label vector $\mathbf{y}_i \in \{0,1\}^C$ satisfies the following conditions:
\begin{equation}
\mathbf{y}_i \preceq \mathbf{y}_i^*, \quad \|\mathbf{y}_i\|_1 = 1,
\end{equation}
where $C$ is the total number of classes and $\preceq$ denotes element-wise no-larger-than comparison. Specifically, for two binary vectors {$\mathbf{a}, \mathbf{b} \in \{0,1\}^C$, $\mathbf{a} \preceq \mathbf{b}$ if $a_j \leq b_j, \forall j$.}

The objective is to train a multi-label classifier $f$ {parametrized by weights $\theta$} to recover the complete set of labels despite incomplete supervision. {The common \ac{AN} strategy} treats all unobserved labels as negatives. This leads to the standard \ac{BCE} loss:
\begin{equation}
\small
\mathcal{L}_{\text{AN}}\left( \theta \right) = - \sum_{i=1}^N \sum_{c=1}^C \left[ y_{i,c} \log p_{i,c} + \left(1 - y_{i,c}\right) \log{\left(1 - p_{i,c}\right)} \right],
\end{equation}
where $\mathbf{p}_i = \texttt{Sigmoid}(f(\mathbf{x}_i; \theta)) \in [0,1]^C$. However, this approach can easily lead to overfitting to label noise by incorrectly penalizing potentially relevant but unobserved labels.

\subsection{Gradient Calibration} \label{sec:meth:gc}

The concept of gradient calibration for handling noisy labels was first proposed in \cite{Liu_2020_NIPS} for {\ac{MCC}}, where \ac{ELR} introduces an auxiliary regularization term to constrain the standard cross-entropy loss, thus preventing the memorization of noisy labels. {The regularization term for {\ac{MCC}}} is defined as follows:
\begin{equation} \label{eq:elr}
    \mathcal{R}_{\text{MCC}}\left( \theta \right) = \frac{1}{n} \sum_{i=1}^{n} \log \left( 1 - \left\langle \mathbf{p}_{i}, \mathbf{t}_{i} \right\rangle \right).
\end{equation}
\cy{Here, both $\mathbf{p}_i$ and $\mathbf{t}_{i}$ are probability vectors whose elements are non-negative and sum to 1. This simplex constraint ensures that their inner product $\left\langle \mathbf{p}_{i}, \mathbf{t}_{i} \right\rangle$ lies within the range $[0,1]$.}
The gradient of \eqref{eq:elr} is defined as follows:
\begin{equation} \label{eq:elr-grad}
    \nabla_\theta\mathcal{R}_{\text{MCC}} \left( \theta \right) = \frac{1}{n} \sum_{i=1}^{n} \nabla_{\theta}f\left(\mathbf{x}_{i}; \theta\right) \cdot \mathbf{g}_{i},
\end{equation}
with $\mathbf{g}_{i} \in \mathbb{R}^C$, whose values are computed as follows:
\begin{equation}
    g_{i,c} = \frac{p_{i,c}}{1 - \langle \mathbf{p}_{i}, \mathbf{t}_{i} \rangle} \sum_{k=1}^{C} \left(t_{i,k} - t_{i,c}\right) \cdot p_{i,k}.
    \label{eq:elr-g}
\end{equation}
As indicated by \eqref{eq:elr-g}, the sign of \eqref{eq:elr-grad} is governed by $t_{i,k} - t_{i,c}$. Specifically, the \ac{ELR} regularization calibrates the gradients toward negative values when $\mathbf{t}_i$ indicates that the class $c$ is the potential correct label. Conversely, if class $c$ is not favored by the pseudo label, the calibration term $g_{i,c}$ remains positive. This behavior is consistent with the gradient direction of the cross-entropy loss, which yields negative gradients for $y_{i,c}=1$ and positive gradients for $y_{i,c}=0$.

\Cref{eq:elr} holds effective only for multi-class cases with $\|\mathbf{p}_i\|_1 = 1$ and $\|\mathbf{t}_i\|_1 = 1$. In the multi-label case, directly applying \eqref{eq:elr} leads to unbounded $\left\langle \mathbf{p}_{i}, \mathbf{t}_{i} \right\rangle$ that also varies significantly across samples due to differing numbers of positive labels. A straightforward adaptation is to extend it to a class-wise binary setting, that is,
\begin{equation} \label{eq:elr-ml-binary}
    \mathcal{R}_{\text{MLC}}^{binary}\left( \theta \right) = \frac{1}{n} \sum_{i=1}^{n}\sum_{c=1}^{C} \log \left( 1 - \left\langle \mathbf{b}_{i,c}, \mathbf{t}_{i,c} \right\rangle \right),
\end{equation}
where $\mathbf{b}_{i,c}=[p_{i,c},1-p_{i,c}]^{\mathsf{T}}$ is the binary prediction vector for sample $i$ at class $c$, and $\mathbf{t}_{i,c}=[t_{i,c},1-t_{i,c}]^{\mathsf{T}}$. In general, negative labels ($y_{i,c}=0$) vastly outnumber positive ones ($y_{i,c}=1$) in multi-label datasets, even though each sample may be associated with multiple classes. Thus, \eqref{eq:elr-ml-binary} falls ineffective in practice, particularly in \ac{SPML} settings, since the regularization term is largely dominated by calibration on $y_{i,c}=0$, and overwhelms the contribution from the sparse positive labels.

To address this limitation, we emphasize the gradient calibration on the potential positive labels with the following regularization term:
\cy{\begin{equation} \label{eq:elr-ml}
    \mathcal{R}_{\text{MLC}}^{\text{GC}}\left( \theta \right) = \frac{1}{n} \sum_{i=1}^{n}\sum_{c=1}^{C}\mathbb{I}(y_{i,c}=0) \log \left( 1 - p_{i,c}\cdot t_{i,c} \right),
\end{equation}
where $\mathbb{I}(\texttt{True})=1$ otherwise 0. Considering the \ac{SPML} setting, \ac{GC} is applied only to negative labels, excluding its use for positive ones in practice. We can compute the gradient of \eqref{eq:elr-ml} as follows:
\begin{equation}
    \nabla_{\theta}\mathcal{R}_{\text{MLC}}^{\text{GC}}\left( \theta \right) = \frac{1}{n} \sum_{i=1}^{n}\sum_{c=1}^{C} \mathbb{I}(y_{i,c}=0)\nabla_{\theta}f_c(\mathbf{x}_{i}; \theta) \cdot g_{i,c},
\end{equation}
where $g_{i,c}$ is defined as follows:
\begin{equation}
    g_{i,c} =  \frac{-t_{i,c}}{1-p_{i,c}\cdot t_{i,c}}\cdot{p_{i,c}(1-p_{i,c})} \leq 0,
\end{equation}
where $p_{i,c}(1-p_{i,c})$ is the sigmoid derivative. We argue that \eqref{eq:elr-ml} is particularly suitable for the \ac{SPML} setting, as it specifically penalizes false negatives by pushing the model to predict positive scores for (negative) labels with higher pseudo-label scores $t_{i,c}$.}
The final objective is a combination of the binary cross-entropy loss and the GC regularization:
\begin{equation} \label{eq:finall}
    \mathcal{L}\left( \theta \right) = \mathcal{L}_{\text{AN}}\left( \theta \right)+\lambda \cdot \mathcal{R}_{\text{MLC}}^{\text{GC}}\left( \theta \right),
\end{equation}
where $\lambda$ is the weight of the regularization term.

However, applying GC from the beginning may lead the model to predict all classes as positive. Liu et al. \cite{liu_aio2_2024} show that adaptively triggering correction before overfitting to label noise is crucial for the effectiveness of sample-correction-related techniques. Inspired by this, in \Cref{sec:meth:earlylearning} we introduce a simple early learning detection mechanism to appropriately activate \ac{GC}. Moreover, as \ac{GC} performance depends on the quality of the pseudo-labels $\mathbf{t}$, in \Cref{sec:meth:pseudolabel} we design a dual-EMA module  to ensure the robust generation of pseudo-labels.

\subsection{Early Learning Detection} \label{sec:meth:earlylearning}
\cy{Pseudo-labelling methods typically require an initial warm-up phase to mitigate the trade-off between additive noise (premature initiation) and subtractive noise overfitting (excessive delay). Ideally, the optimal activation time is identified by the performance plateau on a clean validation set \cite{Arpit_2017_ICML,Zhang_2021_CommACM,liu_aio2_2024}. In \ac{SPML}, where clean validation data is unavailable, we propose monitoring the \ac{mAP} on the noisy validation set. Below, we provide a formal justification for this criterion by analyzing the bounds and stationarity of the noisy metric.}

\subsubsection{Theoretical Framework}
\begin{figure}[t]
    \centering
    \subfigure[Random \ac{SPML}]{\includegraphics[width=0.48\columnwidth]{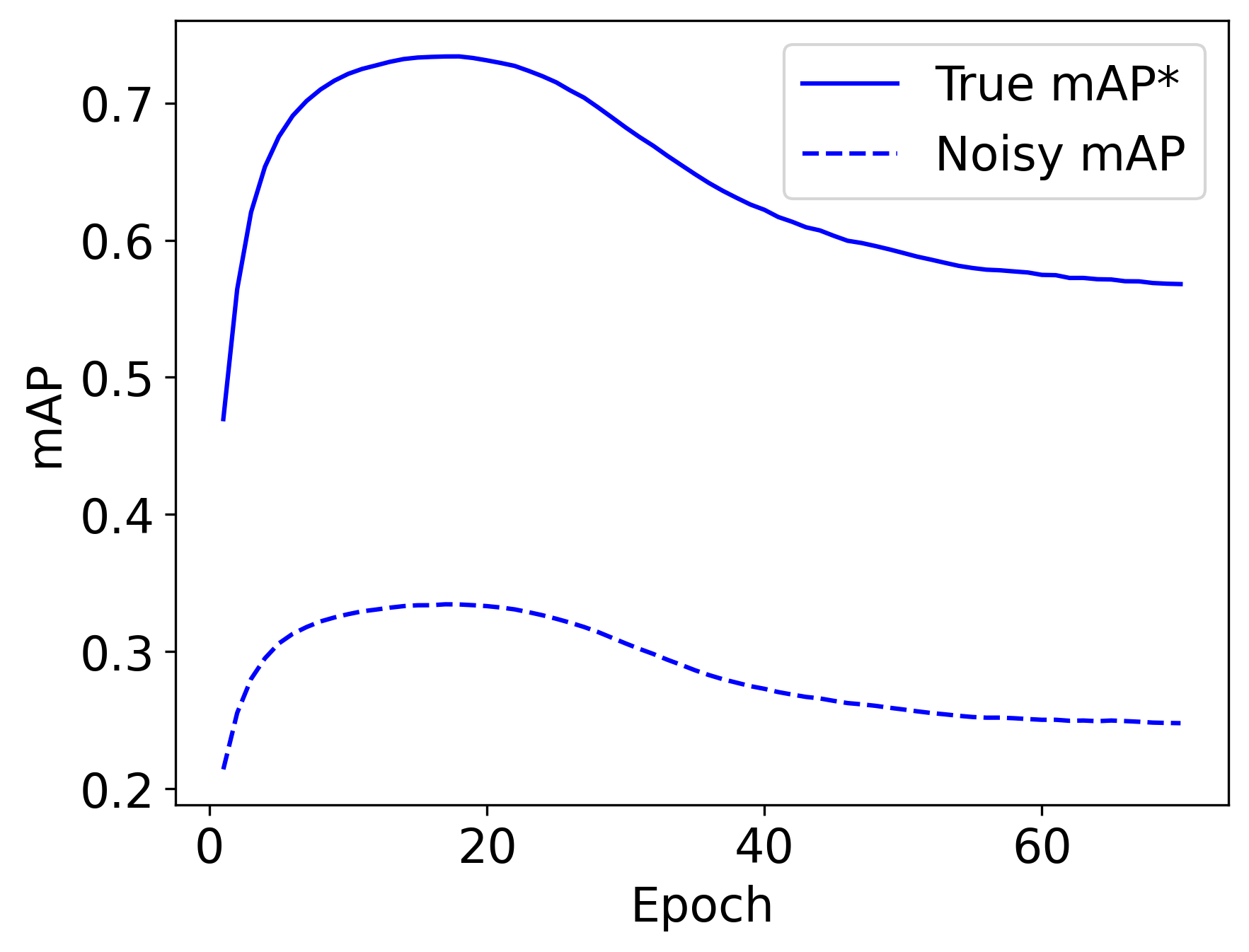}}
    \subfigure[Dominant \ac{SPML}]{\includegraphics[width=0.48\columnwidth]{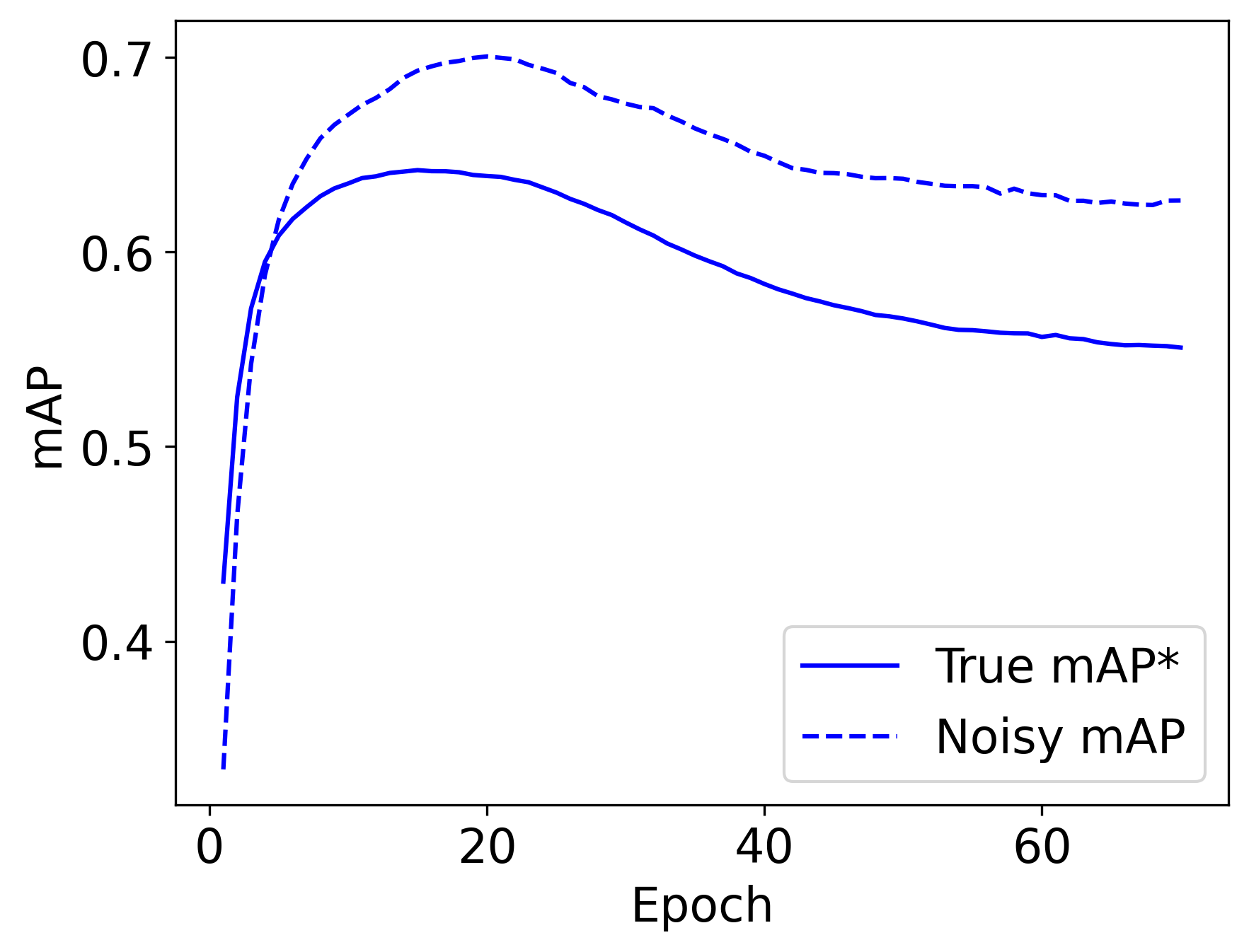}}
    \caption{True and noisy validation \ac{mAP} trends of the teacher model during training on refined BigEarthNet \cite{clasen_reben_2025_arxivonly}: (a) random \ac{SPML}, (b) dominant \ac{SPML}. \cy{The trends align with Propositions 1 and 2, and confirm the validity of the stationarity theorem.}}
    \label{fig:randomiSPML}
\end{figure}
\cy{Let $\text{Rec}^*_c$ and $\text{Prec}^*_c$ denote the clean recall and precision for class $c$. In \ac{SPML}, the set of real positives $\text{P}_c$ is reduced by the flipped positives $\text{F}_c = \beta_c \text{P}_c$, where $\beta_c \in (0,1)$ is the noise rate. Defining $\text{PF}_c = \alpha_c \cdot \text{F}_c$, where $\alpha_c \in [0,1]$,
as the proportion of correctly classified flipped positives, the noisy metrics relate to the clean ones as follows:
\begin{equation} \label{eq:noisy_metrics_general}
\begin{split}
    \text{Rec}_c &=\frac{\text{TP}_c-\text{PF}_c}{\text{P}_c-\text{F}_c} =\frac{\text{Rec}^*_c - \alpha_c \beta_c}{1 - \beta_c}, \\
    \text{Prec}_c &=\frac{\text{TP}_c-\text{PF}_c}{\text{PP}_c} = \text{Prec}^*_c \left( 1-\frac{\alpha_c \beta_c}{\text{Rec}^*_c} \right),
\end{split}
\end{equation}
where $\text{TP}_c$ and $\text{PP}_c$ are the true and predicted positives, respectively.
The \ac{mAP} is the macro average of \ac{AP} for each class:
\begin{equation}
    \text{mAP} = \frac{1}{C}\sum_{c=1}^C{\text{AP}_c},
\end{equation}
where $\text{AP}_c = \int_0^1{\text{Prec}_c\left(\text{Rec}_c\right)d\text{Rec}_c}$.}

\noindent\cy{ \textbf{Proposition 1 (Random \ac{SPML} Lower Bound).} \textit{Under instance-independent noise, the expected noisy \ac{mAP} is a lower bound of the clean $\text{mAP}^*$.}}

\noindent\cy{ \textit{Proof:} In Random \ac{SPML}, the noise is independent of the input features \cite{Perantoni_2022_TGRS}. Thus, the model cannot generalize to the random flips, implying $\mathbb{E}[\alpha_c] \approx \text{Rec}^*_c$. Substituting this into \eqref{eq:noisy_metrics_general} yields $\mathbb{E}[\text{AP}_c] \approx (1 - \beta_c) \text{AP}_c^*$. The expected noisy \ac{mAP} becomes:
\begin{equation} \label{eq:randmap}
    \mathbb{E}[\text{mAP}] \approx (1 - \bar{\beta}) \text{mAP}^* - \text{Cov}(\beta, \text{AP}^*),
\end{equation}
where $\bar{\beta}$ is the mean flip rate. Since better-performing classes generally experience less flipping (negative covariance), $\mathbb{E}[\text{mAP}] < \text{mAP}^*$. \hfill $\square$}

\noindent \cy{Fig.~\ref{fig:randomiSPML}(a) shows the trends observed while training on the refined BigEarthNet \cite{clasen_reben_2025_arxivonly} dataset with random single-positive labels, which are aligned with our theoretical analysis.}

\noindent \cy{\textbf{Proposition 2 (Dominant \ac{SPML} Upper Bound).} \textit{Under instance-dependent dominant noise, the expected noisy \ac{mAP} is an upper bound of the clean $\text{mAP}^*$.}}

\noindent \cy{\textit{Proof:} In Dominant \ac{SPML}, the model learns the label-noise model \cite{Perantoni_2022_TGRS}, ignoring non-dominant true positives ($\mathbb{E}[\alpha_c] \to 0$). Eq. \eqref{eq:noisy_metrics_general} simplifies to $\mathbb{E}[\text{Rec}_c] \approx \frac{\text{Rec}_c^*}{1 - \beta_c}$ and $\mathbb{E}[\text{Prec}_c] \approx \text{Prec}_c^*$, yielding $\mathbb{E}[\text{AP}_c] \approx \frac{1}{1 - \beta_c} \text{AP}_c^*$. The noisy \ac{mAP} becomes:
\begin{equation} \label{eq:domimap}
    \mathbb{E}[\text{mAP}] \approx \overline{\left(\frac{1}{1-\beta}\right)} \text{mAP}^* + \text{Cov}\left(\frac{1}{1-\beta}, \text{AP}^*\right).
\end{equation}
Assuming dominant classes perform better (negative covariance), $\text{mAP}^* < \mathbb{E}[\text{mAP}]$. \hfill $\square$}

\noindent\cy{ Generally, Dominant \ac{SPML} results in three consecutive trends during learning \cite{liu_aio2_2024}: i) an early learning phase where $\mathbb{E} \left[\alpha_c\right] \approx \text{Rec}^*_c$ still holds, yielding similar results to random \ac{SPML}; ii) a transition phase; and iii) a label-noise model fitting phase. Fig.~\ref{fig:randomiSPML}(b) plots the trends observed while training on the refined BigEarthNet \cite{clasen_reben_2025_arxivonly} dataset with dominant single-positive labels, showing this three-phase phenomenon.}

\cy{\subsubsection{Adaptive Triggering Mechanism}
We now address the validity of using the noisy metric for early stopping.}

\noindent\cy{ \textbf{Theorem 1 (Stationarity Preservation).} \textit{The stationary points of the expected noisy \ac{mAP} coincide with those of the clean $\text{mAP}^*$ in the late early-learning phase.}}

\noindent\cy{ \textit{Proof:} Let $t$ denote the training epoch. Combining Propositions 1 and 2, the noisy metric is a linear transformation of the clean metric: $\mathbb{E}[\text{mAP}(t)] = \bar{\omega} \text{mAP}^*(t) + \text{Cov}(\omega, \text{AP}^*(t))$. Taking the temporal derivative:
\begin{equation} \label{eq:derivative}
    \frac{\partial \mathbb{E}[\text{mAP}]}{\partial t} = \bar{\omega} \frac{\partial \text{mAP}^*}{\partial t} + \frac{\partial}{\partial t}\text{Cov}(\omega, \text{AP}^*(t)).
\end{equation}
The noise parameters $\omega$ are constant. 
Near the performance plateau ($t \to t^*$), class-wise learning dynamics tend to stabilize, making the covariance term approximately constant at that time ($\frac{\partial}{\partial t}\text{Cov} \approx 0$)
Consequently:
\begin{equation}
    \frac{\partial \mathbb{E}[\text{mAP}]}{\partial t} \propto \frac{\partial \text{mAP}^*}{\partial t}.
\end{equation}
Thus, identifying the peak of the noisy \ac{mAP} is mathematically equivalent to identifying the peak of the clean \ac{mAP}. \hfill $\square$}

\cy{\textit{Implementation:} To mitigate stochastic fluctuations in the single-run estimates of $\mathbb{E}[\text{mAP}]$, we monitor the \ac{mAP} of the teacher model (maintained via \ac{EMA}), which acts as a low-pass filter on the model's weights (see Fig.~\ref{fig:exp:abla:AN}). We select the epoch maximizing the teacher's noisy validation \ac{mAP} (with a patience of $b_e$ epochs) to initiate pseudo-labeling.}

\subsection{Dual-EMA Pseudo-Label Generation} \label{sec:meth:pseudolabel}

As discussed in \Cref{sec:meth:gc}, \ac{GC} relies on high-quality pseudo-labels. The teacher’s predictions are a natural choice, since the early learning detection is on the teacher model.
However, due to slow weight updates, the teacher model may exhibit a delay in incorporating recently acquired knowledge after \ac{GC} begins. Therefore, we propose to combine predictions from both teacher and student models. Specifically, we apply \ac{EMA} to the student’s predictions with a smaller smoothing factor than that for the teacher model, to avoid overfitting to its own outputs while maintaining adaptability.

Let $\mathbf{p}_i^T$ and $\mathbf{p}_i^S$ denote the prediction probabilities of the teacher and student models for sample $i$, respectively. For the teacher model, we have the following:
\begin{equation}
\mathbf{p}_i^T(e) = f(\mathbf{x}_i;\theta^{T}(e)),
\label{eq:ema-teacher-prediction}
\end{equation}
\begin{equation}
    \theta^{T}(e)=\beta_t \cdot \theta^{T}(e-1)+(1-\beta_t)\cdot \theta^{S}(e)
\label{eq:ema-teacher-weight}
\end{equation}
with $\theta^{T}(e)$ and $\theta^{S}(e)$ being the model weights of the teacher and student models at iteration $e$. We set the smoothing coefficient $\beta_t=0.999$ following \cite{tarvainen_mean_2017}. 
Then, the temporarily \ac{EMA}-smoothed student predictions are as follows:
\begin{equation}
\tilde{\mathbf{p}}_i^S(e) = \beta_s \cdot \tilde{\mathbf{p}}_i^S(e-1) + (1 - \beta_s) \cdot \mathbf{p}_i^S(e).
\label{eq:ema-student}
\end{equation}
We empirically set $\beta_s=0.8$ to guarantee the update speed of the student's predictions. Then, the pseudo-label $\mathbf{t}_i(e)$ used for gradient calibration is computed as the combination of the teacher and student predictions as follows:
\begin{equation}
\mathbf{t}_i = \gamma \cdot \mathbf{p}_i^T + (1-\gamma) \cdot \tilde{\mathbf{p}}_i^S,
\label{eq:pseudo-label}
\end{equation}
where $e$ is omitted for simplicity. In our experiments, we set $\gamma = 0.5$ to balance the contributions of the teacher and student models when generating pseudo-labels.

\subsection{Mixup} \label{sec:meth:mixup}
To further enhance robustness against label noise and improve generalization, we incorporate Mixup \cite{Zhang_2018_ICLR} into the \ac{GC} stage. For each batch, we randomly sample a pairing input $(\mathbf{x}_j, \mathbf{y}_j, \mathbf{t}_j)$ for every original input $(\mathbf{x}_i, \mathbf{y}_i, \mathbf{t}_i)$, and compute:
\begin{equation}
\tilde{\mathbf{z}}_i = \phi \cdot \mathbf{z}_i + (1 - \phi) \cdot \mathbf{z}_j \quad \text{for} \quad \mathbf{z} \in \{\mathbf{x}, \mathbf{y}, \mathbf{t}\},
\label{eq:mixup}
\end{equation}
where $\phi \sim \text{Beta}(\alpha, \alpha)$ and $\alpha > 0$ controls the interpolation strength. The mixed samples $(\tilde{\mathbf{x}}_i, \tilde{\mathbf{y}}_i,\tilde{\mathbf{t}}_i)$ are then used for training during the \ac{GC} phase. This data augmentation encourages linear behavior between examples, mitigates overfitting, and stabilizes training under noisy labels.

\subsection{\cy{Complexity Analysis}} \label{sec:meth:compl}
\cy{To assess practical feasibility, we analyze the computational complexity of \ac{AdaGC} relative to a standard classification baseline. First, one can note that \ac{AdaGC} is a training-only framework, \textit{i.e.}, the dual-EMA module is discarded after training. Consequently, the inference complexity (parameters and FLOPs) is identical to the student backbone, ensuring no additional latency or memory overhead during deployment.}

\cy{Regarding training efficiency, a standard iteration incurs a cost of approximately $3\times$ the forward pass FLOPs ($1\text{F}$ forward and $\sim2\text{F}$ backward). Instead, \ac{AdaGC} processes two views during the \ac{GC} stage: the original batch and a Mixup-augmented batch. However, the original batch is utilized exclusively for generating robust pseudo-labels via the student and teacher networks, which requires only forward passes, while the model optimization relies on the Mixup batch. Therefore, the computationally intensive backward pass is strictly necessary only for this view. This design results in a total theoretical cost of $\sim5\text{F}$ per iteration, representing a moderate $\sim66\%$ overhead compared to the baseline ($3\text{F}$) exclusively during the \ac{GC} stage. Note that Mixup is simply a linear combination of two images. Thus, its additional FLOPs are mathematically negligible compared to the rest of the pipeline.}

\cy{In terms of memory complexity, the overhead is modest. The teacher \ac{EMA} maintains only a second copy of the backbone parameters (not activations), contributing less than 10\% of total memory in typical large-scale models. The temporally smoothed student predictions are stored as image-wise labels, which also require minimal memory. Taken together, both FLOPs and memory considerations indicate that \ac{AdaGC} is computationally feasible for large-scale remote sensing applications.}

\begin{table}[htp]
    \centering
    \renewcommand{\arraystretch}{1.3} 
    \setlength\dashlinedash{2pt}
    \setlength\dashlinegap{2pt}
    
    \caption{Refined BigEarthNet Land Cover Classes and Related Flip\\
    Rates $\beta_c$ on the Training Subset in the Two Cases of\\
    Random and Dominant \ac{SPML} Label Noise
    }
    \label{tab:ben-stats}
    
    \begin{tabularx}{\columnwidth}{@{} X c c r @{}}
        \hline\hline 
        \textbf{Class Name} & \textbf{Random $\beta_c$} & \textbf{Dominant $\beta_c$} & \textbf{Support} \\
        \hline 
        
        Urban fabric & 0.73 & 0.84 & 33,626 \\ \hdashline
        Industrial or commercial units & 0.72 & 0.91 & 6,398 \\ \hdashline
        Arable land & 0.63 & 0.51 & 88,236 \\ \hdashline
        Permanent crops & 0.70 & 0.74 & 15,190 \\ \hdashline
        Pastures & 0.66 & 0.62 & 42,968 \\ \hdashline
        Complex cultivation patterns & 0.73 & 0.77 & 51,034 \\ \hdashline
        Land principally occupied by agriculture, with significant areas of natural vegetation & 0.73 & 0.86 & 60,127 \\ \hdashline
        Agro-forestry areas & 0.64 & 0.54 & 15,082 \\ \hdashline
        Broad-leaved forest & 0.69 & 0.65 & 64,164 \\ \hdashline
        Coniferous forest & 0.66 & 0.54 & 76,344 \\ \hdashline
        Mixed forest & 0.70 & 0.67 & 78,791 \\ \hdashline
        Natural grassland and sparsely vegetated areas & 0.69 & 0.82 & 6,984 \\ \hdashline
        Moors, heathland and sclerophyllous vegetation & 0.70 & 0.70 & 6,329 \\ \hdashline
        Transitional woodland, shrub & 0.70 & 0.85 & 64,198 \\ \hdashline
        Beaches, dunes, sands & 0.72 & 0.94 & 738 \\ \hdashline
        Inland wetlands & 0.67 & 0.78 & 11,344 \\ \hdashline
        Coastal wetlands & 0.70 & 0.70 & 670 \\ \hdashline
        Inland waters & 0.67 & 0.73 & 30,466 \\ \hdashline
        Marine waters & 0.08 & 0.05 & 35,914 \\ 
        \hline 
        
        \textbf{Micro average} & \textbf{0.65} & \textbf{0.65} & -- \\ \hdashline
        \textbf{Macro average $\bar{\beta}$} & \textbf{0.66} & \textbf{0.70} & -- \\ 
        \hline 
        
        \multicolumn{3}{l}{Average number of true positive labels per image} & 2.9 \\
        \hline\hline 
    \end{tabularx}
\end{table}
\section{Experiments} \label{sec:experiments}
This section presents the experimental setups, comparative results, and a comprehensive ablation study of AdaGC.

\subsection{Datasets}  \label{sec:experiments:data}

We evaluate the proposed method on two multi-label \ac{RS} benchmark datasets with different modalities and resolutions: \ac{reBEN} \cite{Sumbul_ben_2019_IGARSS, sumbul_BigEarthNet-mm_2021, clasen_reben_2025_arxivonly}, derived from SAR and multispectral Sentinel-1/2 imagery at 10m resolution, and AID-multilabel \cite{hua_relation_2020}, derived from high-resolution RGB images at 0.5–8m resolution. While \ac{reBEN} focuses on land cover classification, AID-multilabel includes both land cover categories and object-level semantic information.

\subsubsection{reBEN}

\begin{table}[ht]
    \centering
    \renewcommand{\arraystretch}{1.3}
    \setlength\dashlinedash{2pt}
    \setlength\dashlinegap{2pt}
    
    \caption{
    AID Multilabel Classes and Related Flip Rates $\beta_c$ on \\
    the Training Subset in the Two Cases of Random \\
    and Manual \ac{SPML} Label Noise
    }
    \label{tab:aid-stats}
    
    \begin{tabularx}{\columnwidth}{@{} X c c r @{}} 
        \hline\hline 
        \textbf{Class Name} & \textbf{Random $\beta_c$} & \textbf{Manual $\beta_c$} & \textbf{Support} \\
        \hline
        
        Airplane  & 0.83 & 0.80 & 59 \\ \hdashline
        Bare-soil & 0.80 & 0.78 & 882 \\ \hdashline
        Buildings & 0.82 & 0.81 & 1,316 \\ \hdashline
        Cars      & 0.85 & 0.85 & 1,210 \\ \hdashline
        Chaparral & 0.76 & 0.85 & 79 \\ \hdashline
        Court     & 0.86 & 0.78 & 203 \\ \hdashline
        Dock      & 0.94 & 0.83 & 166 \\ \hdashline
        Field     & 0.66 & 0.57 & 127 \\ \hdashline
        Grass     & 0.78 & 0.80 & 1,368 \\ \hdashline
        Pavement  & 0.83 & 0.84 & 1,367 \\ \hdashline
        Sand      & 0.41 & 0.48 & 153 \\ \hdashline
        Sea       & 0.77 & 0.67 & 133 \\ \hdashline
        Ship      & 0.90 & 0.89 & 181 \\ \hdashline
        Tanks     & 0.82 & 0.73 & 66 \\ \hdashline
        Trees     & 0.79 & 0.80 & 1,467 \\ \hdashline
        Water     & 0.83 & 0.84 & 510 \\ 
        \hline 
        
        \textbf{Micro average} & \textbf{0.81} & \textbf{0.81} & -- \\ \hdashline
        \textbf{Macro average $\bar{\beta}$} & \textbf{0.79} & \textbf{0.77} & -- \\ 
        \hline
        
        \multicolumn{3}{l}{Average number of true positive labels per image} & 5.2 \\
        \hline\hline 
    \end{tabularx}
\end{table}
It contains 549,488 Sentinel-1/2 image pairs of size of $120\times120$ pixels covering 10 European countries, each aligned with a pixel-level \ac{CLC} reference map. Multi-label annotations are derived from these maps, resulting in 19 land cover classes (see Table \ref{tab:ben-stats}). Following the splits in \cite{clasen_reben_2025_arxivonly}, we use a 2:1:1 ratio for training, validation, and testing. Each input has 14 channels by concatenating 2 Sentinel-1 and 12 Sentinel-2 L2A bands. Leveraging the reference maps, on this dataset we simulate both types of \ac{SPML} label noise described in \Cref{sec:meth:earlylearning} (\textit{Random} and \textit{Dominant}). \cy{The \textit{Dominant} case better reflects real-world single-positive annotations, as low-resolution images contain limited details, where the prevailing class is more likely to be observed while minor classes are ignored.} Dataset statistics for the two cases are provided in Table \ref{tab:ben-stats}.
\subsubsection{AID-multilabel}
It extends the AID scene classification dataset \cite{xia_aid_2017} with multi-label annotations. The dataset comprises 3,000 high-resolution RGB aerial image patches ($600 \times 600$ pixels) across 17 classes, collected from China, U.S., and several European countries. We exclude the extremely rare “mobile home” class (only 2 samples), resulting in 16 classes. Following \cite{hua_relation_2020}, 20\% of the data is held out for testing. We then use 25\% of the remaining training set for validation.
\cy{Unlike \ac{reBEN}, AID’s high spatial resolution allows multiple objects to be easily recognized, making it difficult to identify a single dominant class. Therefore, in addition to \textit{Random} simulation, we manually annotated the AID dataset in a single-positive manner to assess model performance under realistic SPML conditions. For each image, four randomly shuffled candidate classes were displayed at a time to speed up annotation, with the option to refresh if none were suitable. Consecutive images were also prevented from being assigned to the same class to reduce potential annotation bias. This procedure makes the \textit{Manual} labels behave similarly to random SPML labels, yet retain a slight human selection bias. The entire annotation process took approximately 4.5 hours. We also found and corrected several label inconsistencies in the original AID-multilabel data, most of which are from the meadow folder. The corrected multi-label annotations, the manual single-positive labels, and the annotation tool will be released publicly along with the AdaGC code. Dataset statistics are provided in Table \ref{tab:aid-stats}.} 
For simplicity, in the following we use AID to refer to AID-multilabel.

\begin{table*}[ht]
    \centering
    \caption{Test Performance Comparison of Different Methods on the reBEN-Random Dataset. The Best Average Metric Values Are Reported in Bold, Second Bests Are in Italic. The Related Standard Deviations Are Reported in Brackets}
    \label{tab:ben-rand-results}
    \resizebox{\textwidth}{!}{%
    \begin{tblr}{
      colspec = {llccccccc}, 
      cell{2}{1} = {r=2}{},
      cell{4}{1} = {r=4}{},
      cell{8}{1} = {r=5}{},
      cell{14}{1} = {r=4}{},
      cell{18}{1} = {r=2}{},
      cell{21}{1} = {c=9}{},
      hline{1,21} = {-}{},
      hline{1,21} = {2}{-}{},
      hline{2} = {},
      hline{4} = {-}{},
      hline{4} = {2}{-}{},
      hline{8,13-14,18,20} = {dashed},
      cell{20}{3} = {font=\bfseries,Alto},
      cell{20}{4} = {font=\bfseries,Alto},
      cell{20}{5} = {font=\bfseries,Alto},
      cell{20}{6} = {font=\bfseries,Alto},
      cell{20}{7} = {font=\bfseries,Alto},
      cell{8}{3}  = {font=\itshape,Alto},
      cell{18}{4} = {font=\itshape,Alto},
      cell{18}{5} = {font=\itshape,Alto},
      cell{10}{6} = {font=\itshape,Alto},
      cell{8}{7}  = {font=\itshape,Alto},
    }
                                &           & \textbf{mAP} (\%) $\uparrow$     & \textbf{Coverage} $\downarrow$      & \textbf{Rankloss} (\%) $\downarrow$  & \textbf{OA} (\%) $\uparrow$       & \textbf{mF1} (\%) $\uparrow$     & \textbf{mprecision}  (\%) $\uparrow$   & \textbf{mrecall} (\%) $\uparrow$ \\
Ground-truth Training           & GT w/ BCE & 71.75 (0.04) & 4.264 (0.003) & 3.428 (0.005) & 93.09 (0.01) & 66.63 (0.12) & 72.07 (0.02) & 63.24 (0.35) \\
                                & IUN       & 70.46 (0.11) & 4.331 (0.009) & 3.608 (0.024) & 91.98 (0.01) & 56.00 (0.19) & 84.75 (0.55) & 44.66 (0.18) \\
Baseline Methods                & AN        & 54.97 (0.12) & 4.849 (0.002) & 5.847 (0.003) & 87.13 (0.04) & 28.15 (0.30) & 72.93 (1.67) & 19.00 (0.21) \\
                                & AN-LS     & 54.40 (0.66) & 5.375 (0.114) & 7.051 (0.309) & 86.90 (0.03) & 25.47 (0.44) & 75.90 (0.73) & 17.07 (0.31) \\
                                & WAN       & 59.43 (0.34) & 4.660 (0.010) & 5.089 (0.037) & 90.16 (0.05) & 60.15 (0.20) & 55.96 (0.36) & 66.14 (0.65) \\
                                & EPR\textsuperscript{a}       & 50.74 (0.40) & 5.206 (0.022) & 6.825 (0.053) & 80.13 (0.08) & 51.22 (0.11) & 38.25 (0.13) & 85.59 (0.19) \\
General Label Noise Methods     & ELR       & 62.40 (0.08) & 4.860 (0.021) & 4.890 (0.021) & 91.45 (0.04) & 63.97 (0.07) & 62.63 (0.36) & 68.26 (0.25) \\
                                & LL-Cp     & 57.27 (0.13) & 4.983 (0.013) & 6.068 (0.038) & 89.60 (0.03) & 49.35 (0.58) & 65.47 (0.40) & 43.67 (0.88) \\
                                & LL-Ct     & 62.35 (0.23) & 4.722 (0.010) & 4.856 (0.029) & 91.76 (0.03) & 54.87 (0.35) & 76.27 (1.09) & 46.41 (0.18) \\
                                & LL-R      & 59.70 (0.27) & 4.688 (0.012) & 5.054 (0.042) & 90.85 (0.05) & 53.91 (0.56) & 70.44 (1.08) & 45.45 (0.45) \\
                                & BoostLU   & 61.77 (0.63) & 4.662 (0.011) & 4.780 (0.023) & 91.59 (0.01) & 55.41 (0.61) & 74.80 (1.30) & 46.32 (0.36) \\
PUL Methods                     & Dist-PU\textsuperscript{a}   & 61.98 (0.26) & 4.860 (0.016) & 5.111 (0.024) & 90.67 (0.04) & 61.49 (0.35) & 59.11 (0.47) & 68.83 (0.52) \\
SPML Methods                    & ROLE      & 55.73 (0.36) & 6.479 (0.007) & 6.424 (0.020) & 91.18 (0.05) & 59.65 (0.31) & 61.90 (0.45) & 58.36 (0.35) \\
                                & GR        & 57.38 (0.49) & 4.914 (0.034) & 5.855 (0.120) & 89.77 (0.08) & 56.79 (0.85) & 57.91 (0.16) & 56.06 (1.56) \\
                                & LAC       & 57.88 (0.30) & 4.801 (0.021) & 5.384 (0.036) & 87.33 (0.25) & 28.97 (1.38) & 72.29 (0.62) & 19.85 (1.11) \\
                                & MIME      & 58.64 (0.35) & 4.894 (0.017) & 5.632 (0.064) & 90.66 (0.13) & 55.13 (0.36) & 62.40 (0.46) & 50.39 (0.21) \\
\cy{RS-specific Methods}        & \cy{RCML} & 64.91 (0.12) & 4.585 (0.005) & 4.467 (0.020) & 91.24 (0.04) & 52.90 (0.32) & 78.57 (1.29) & 42.42 (0.34) \\
                                & \cy{LCR}  & 54.56 (0.32) & 4.864 (0.009) & 5.912 (0.044) & 87.16 (0.03) & 28.52 (0.14) & 73.82 (1.81) & 19.31 (0.08) \\
Proposed                        & AdaGC     & 67.85 (0.05) & 4.553 (0.012) & 4.165 (0.033) & 92.49 (0.01) & 64.46 (0.24) & 70.43 (0.59) & 63.22 (0.10) \\
\textsuperscript{a} EPR and Dist-PU use additional a-priori information, which is assumed unknown in all other approaches.
    \end{tblr}}
\end{table*}

\subsection{Experimental Setup}  \label{sec:experiments:setting}

We adopt ResNet-34 and ResNet-50 as backbones for the \ac{reBEN} and AID datasets, with batch sizes of 512 and 32, respectively. ResNet-34 is chosen for reBEN as an efficient compromise, since our focus is on evaluating method effectiveness rather than maximizing absolute accuracy. 
For AID, we use ImageNet-pretrained weights given its RGB input. \cy{To adhere to a realistic weakly supervised scenario where ground-truth multi-labels are unavailable, we strictly avoid the use of clean data for model selection. Consequently, all hyperparameters are tuned via grid search on the noisy validation set, simulating the absence of clean labels. While this strategy relies on imperfect supervision, it prevents the introduction of ``\textit{oracle}'' information and ensures that the reported performance reflects the method's capability to generalize without privileged access to clean labels. Despite the use of noisy validation labels, the resulting metrics remain informative for model selection. Under \ac{SPML} noise, random errors are inherently difficult for deep models to memorize, and thus the validation signal largely reflects the underlying clean structure. As confirmed in our experiments, noisy \ac{mAP} validation scores maintain a strong correlation with clean test performance, making them a practical and reliable indicator in the absence of ground-truth labels.} 
To reduce computational cost, tuning is performed on subsets (20\% for reBEN, 50\% for AID) and fewer epochs (10) for comparison methods. For \ac{AdaGC}, the Mixup Beta parameter $\alpha$ is set to 1. \cy{The \ac{GC} loss weight $\lambda$ is set to 3, 5, 9, and 8 for \ac{reBEN}-\textit{Random}, \ac{reBEN}-\textit{Dominant}, AID-\textit{Random} and AID-\textit{Manual}, respectively, and the learning rate is set to 1e-3 and 5e-4 for \ac{reBEN} and AID, respectively, based on tuning results.}

For the final evaluation, all models are trained for 70 epochs on the full training sets using the optimal hyperparameters and evaluated on the \cy{clean} test sets. To ensure robustness and statistical reliability, each experiment is repeated for three Monte Carlo runs. We report the mean and unbiased standard deviation of all metrics.

\begin{table*}
    \centering
    \caption{Test Performance Comparison of Different Methods on the reBEN-Dominant Dataset. The Best Average Metric Values Are Reported in Bold, Second Bests Are in Italic. The Related Standard Deviations Are Reported in Brackets}
    \label{tab:ben-domi-results}
    \resizebox{\textwidth}{!}{%
    \begin{tblr}{
      colspec = {llccccccc}, 
      cell{2}{1} = {r=2}{},
      cell{4}{1} = {r=4}{},
      cell{8}{1} = {r=5}{},
      cell{14}{1} = {r=4}{},
      cell{18}{1} = {r=2}{},
      cell{21}{1} = {c=9}{},
      hline{1,21} = {-}{},
      hline{1,21} = {2}{-}{},
      hline{2} = {},
      hline{4} = {-}{},
      hline{4} = {2}{-}{},
      hline{8,13-14,18,20} = {dashed},
      cell{20}{3} = {font=\bfseries,Alto},
      cell{20}{4} = {font=\bfseries,Alto},
      cell{20}{5} = {font=\bfseries,Alto},
      cell{8}{6}  = {font=\bfseries,Alto},
      cell{20}{7} = {font=\bfseries,Alto},
      cell{6}{3}  = {font=\itshape,Alto},
      cell{15}{4} = {font=\itshape,Alto},
      cell{15}{5} = {font=\itshape,Alto},
      cell{11}{6} = {font=\itshape,Alto},
      cell{12}{7} = {font=\itshape,Alto},
    }
                                &           & \textbf{mAP} (\%) $\uparrow$     & \textbf{Coverage} $\downarrow$      & \textbf{Rankloss} (\%) $\downarrow$ & \textbf{OA} (\%) $\uparrow$       & \textbf{mF1} (\%) $\uparrow$     & \textbf{mprecision}  (\%) $\uparrow$   & \textbf{mrecall} (\%) $\uparrow$ \\
Ground-truth Training           & GT w/ BCE & 71.75 (0.04) & 4.264 (0.003)  &  3.428 (0.005) & 93.09 (0.01) & 66.63 (0.12) & 72.07 (0.02) & 63.24 (0.35) \\
                                & IUN       & 64.46 (0.07) & 5.289 (0.013)  &  5.908 (0.007) & 90.78 (0.01) & 47.38 (0.09) & 86.11 (0.95) & 35.73 (0.06) \\
Baseline Methods                & AN        & 55.91 (0.29) & 6.533 (0.075)  &  8.128 (0.084) & 88.70 (0.01) & 34.25 (0.26) & 84.35 (0.49) & 24.08 (0.15) \\
                                & AN-LS     & 53.65 (0.12) & 8.474 (0.088)  & 14.878 (0.282) & 88.61 (0.00) & 33.93 (0.16) & 85.99 (0.99) & 23.75 (0.12) \\
                                & WAN       & 58.69 (0.28) & 6.627 (0.054)  &  8.153 (0.044) & 90.76 (0.02) & 50.56 (0.27) & 74.22 (1.12) & 41.50 (0.14) \\
                                & EPR\textsuperscript{a}       & 49.15 (0.36) & 6.150 (0.034)  &  9.003 (0.107) & 87.22 (0.07) & 53.34 (0.20) & 48.48 (0.47) & 64.38 (0.39) \\
General Label Noise Methods     & ELR       & 54.48 (0.33) & 9.149 (0.015)  & 11.429 (0.087) & 91.02 (0.02) & 50.08 (0.10) & 76.38 (0.81) & 42.05 (0.13) \\
                                & LL-Cp     & 47.70 (0.20) & 6.061 (0.016)  &  8.968 (0.012) & 88.04 (0.04) & 53.17 (0.11) & 51.47 (0.49) & 59.48 (0.30) \\
                                & LL-Ct     & 54.11 (0.27) & 6.287 (0.107)  &  8.014 (0.216) & 90.88 (0.14) & 53.67 (0.12) & 70.67 (0.90) & 47.82 (0.25) \\
                                & LL-R      & 54.41 (0.13) & 7.369 (0.081)  &  8.662 (0.055) & 91.01 (0.02) & 53.59 (0.20) & 70.83 (0.96) & 46.88 (0.16) \\
                                & BoostLU   & 47.36 (0.34) & 6.711 (0.110)  &  9.767 (0.376) & 88.62 (0.21) & 54.70 (0.45) & 57.40 (1.38) & 58.96 (0.46) \\
PUL Methods                     & Dist-PU\textsuperscript{a}   & 53.51 (0.32) & 6.224 (0.109)  &  8.770 (0.260) & 88.23 (0.09) & 53.17 (0.29) & 56.21 (1.78) & 60.53 (0.17) \\
SPML Methods                    & ROLE      & 48.54 (0.41) & 12.465 (0.083) & 14.591 (0.157) & 90.81 (0.03) & 50.24 (0.46) & 75.38 (1.05) & 40.88 (0.47) \\
                                & GR        & 55.73 (0.23) & 5.765 (0.050)  &  7.238 (0.097) & 89.20 (0.00) & 38.28 (0.18) & 81.43 (0.56) & 27.57 (0.16) \\
                                & LAC       & 54.67 (0.56) & 6.029 (0.074)  &  8.170 (0.206) & 88.57 (0.02) & 33.43 (0.63) & 80.57 (1.58) & 23.67 (0.37) \\
                                & MIME      & 55.61 (0.07) & 6.417 (0.061)  &  8.982 (0.125) & 89.90 (0.03) & 43.54 (0.39) & 76.38 (1.19) & 33.36 (0.55) \\
\cy{RS-specific Methods}        & \cy{RCML} & 58.26 (0.82) & 5.890 (0.134)  &  8.035 (1.181) & 88.73 (1.05) & 40.75 (1.27) & 83.06 (2.35) & 30.93 (2.88) \\
                                & \cy{LCR}  & 55.64 (0.20) & 6.579 (0.045)  &  8.151 (0.080) & 88.67 (0.00) & 34.25 (0.06) & 83.24 (0.54) & 24.05 (0.04) \\
Proposed                        & AdaGC     & 60.06 (0.16) & 5.581 (0.026)  &  6.877 (0.054) & 90.00 (0.06) & 59.08 (0.25) & 60.19 (0.47) & 63.43 (0.37)  \\ 
\textsuperscript{a} EPR and Dist-PU use additional a-priori information, which is assumed unknown in all other approaches.
    \end{tblr}}
\end{table*}

In addition to \ac{mAP}, we report two ranking-based metrics, \textit{i.e.}, coverage and ranking loss (Rankloss), and four accuracy-based metrics, \textit{i.e.}, \ac{OA}, \ac{mF1}, mean precision (mprecision), and mean recall (mrecall). The metrics are defined as follows.
\begin{itemize}
    \item \textbf{Coverage} measures how far we need to go down the ranked label list to cover all true labels:
        \begin{equation}
        \frac{1}{N} \sum_{i=1}^{N} \max_{y \in \mathbf{y}_i} \text{rank}_f(\mathbf{x}_i, y) - 1,
        \end{equation}
        where $\text{rank}_f(x_i, y)$ is the position of the \ac{GT} label $y_{i,c}$ in the ranking list predicted by $f$;
    \item \textbf{Ranking loss} computes the average fraction of label pairs that are incorrectly ordered:
        \begin{equation} 
        \small
        \frac{1}{N} \sum_{i=1}^{N} \frac{1}{|\mathbf{y}_i||\overline{\mathbf{y}_i}|} \sum_{(y_{1}, y_{0}) \in \mathbf{y}_i \times \overline{\mathbf{y}_i}} \mathbb{I}(f(\mathbf{x}_i, y_{1}) \leq f(\mathbf{x}_i, y_{0})),
        \end{equation}
        where $\mathbf{y}_i$ and $\overline{\mathbf{y}_i}$ are the sets of relevant (positive) and irrelevant (negative) labels, respectively, and $\mathbb{I}(\texttt{True})=1$ otherwise 0;
    \item \textbf{\ac{OA}} measures the percentage of samples with all labels correctly predicted:
        \begin{equation}
        \frac{1}{N} \sum_{i=1}^{N}\sum_{c=1}^{C} \mathbb{I}(\hat{\mathbf{y}}_{i,c} = \mathbf{y}_{i,c});
        \end{equation}
    \item \textbf{\ac{mF1}} is the average of F1-scores computed per class:
        \begin{equation}
        \frac{1}{C} \sum_{c=1}^{C} \frac{2 \cdot \text{Prec}_c \cdot \text{Rec}_c}{\text{Prec}_c + \text{Rec}_c}.
        \end{equation}
\end{itemize} 

We compared AdaGC against a wide array of existing methods, categorized as follows.
\begin{itemize}
    \item \textbf{Baseline methods}: We considered several baselines \cite{Cole_2021_CVPR}, such as \ac{AN}, \ac{AN-LS}, \ac{WAN}, and \ac{EPR}. All use a \ac{BCE} loss at their core, while \ac{EPR} uses only the \ac{BCE} positive term and adds a regularization term, assuming the a-priori knowledge of the average number of positive labels per sample (see Tables \ref{tab:ben-stats} and \ref{tab:aid-stats}).
    \item \textbf{General label noise methods}: From this category, we considered the \ac{ELR} approach adapted as in \cite{Burgert_2022_TGRS}, the \ac{LL} family of approaches (\textit{i.e.}, LL-R, LL-Ct, and LL-Cp) \cite{Kim_2022_CVPR}, and the BoostLU \cite{Kim_2023_CVPR} in combination with LL-Ct.
    \item \textbf{\ac{PUL} methods}: We considered a single \ac{SOTA} method from this category, \textit{i.e.}, Dist-PU \cite{Zhao_2022_CVPR}, adapted for \ac{SPML} by simply applying it to each class independently.
    \item \textbf{\ac{SPML}-specific methods}: The main comparison methods we considered for \ac{SPML} are \ac{ROLE} \cite{Cole_2021_CVPR}, \ac{GR Loss} \cite{Chen_2024_IJCAI}, \ac{LAGC} \cite{Xie_2022_NIPS}, and \ac{MIME} \cite{Liu_2023_ICML}.
    \item \textbf{Upper bound methods}: For analytical comparison with \ac{GT} training, we considered two cases: i) training with  \ac{GT} multi-labels using a \ac{BCE} loss (GT w/ BCE), and ii) the \ac{IUN} approach \cite{Cole_2021_CVPR}, where in addition to the single positive labels, all true negatives are provided for training. \ac{IUN} allows us to assess the performance in a hypothetical scenario where all true negatives have been retrieved. False negatives are ignored during training, meaning that the model is still provided with a single positive label for each image.
    \item \textbf{RS-specific methods}: \cy{We also included RCML \cite{Aksoy_2024_TNNLS} and LCR \cite{hua_learning_2020}, which were originally developed for general noisy multi-label RS classification. RCML leverages a collaborative learning framework with a ranking loss for sample selection to mitigate label noise, while LCR uses word embedding similarity to regularize model training under noisy annotations. To adapt them to SPML, we use only the missing-class labels term in RCML and perform selection per class rather than per sample. For LCR, we use RemoteCLIP's \cite{liu_remoteclip_2024} text encoder to encode multi-word class names. These adaptations ensure a fair comparison while preserving the core noise-robust strategies of the original methods.}
\end{itemize}

\subsection{Benchmark Results} \label{sec:experiments:benchmark}

This section analyzes the benchmark results reported in Tables \ref{tab:ben-rand-results}–\ref{tab:aid-results-manual}. Across all scenarios, \ac{AdaGC} consistently achieves the highest \ac{mAP} and strong \ac{SPML} performance in terms of coverage, rankloss, and \ac{mF1}, demonstrating its effectiveness and robustness in handling \ac{SPML} in \ac{RS} imagery. The upper-bound \ac{IUN} performs well in \ac{mAP} but suffers from low mrecall due to its inherent single-positive bias. \ac{AN} and \ac{AN-LS} are highly sensitive to false negatives, exhibiting high mprecision but low mrecall and consequently poor \ac{mAP}. In contrast, \ac{WAN} is more robust and achieves the best baseline \ac{mAP} by better balancing precision-recall trade-offs. \ac{EPR}, trained exclusively on positive labels with a cardinality regularizer, consistently shows the highest mrecall across datasets but low mprecision, reflecting its tendency to overestimate positives. Its advantage largely stems from the use of the true average number of labels per image, a prior not shared by other competitors. \cy{The two RS-specific methods designed for general label noise exhibit clear limitations under \ac{SPML} conditions.} The following subsections detail dataset- and scenario-specific observations.

\subsubsection{\ac{reBEN}-\textit{Random}}

\begin{table*}
    \centering
    \caption{Test Performance Comparison of Different Methods on the \cy{AID-Random} Dataset. The Best Average Metric Values Are Reported in Bold, Second Bests Are in Italic. The Related Standard Deviations Are Reported in Brackets}
    \label{tab:aid-results}
    \resizebox{\textwidth}{!}{%
    \begin{tblr}{
      colspec = {llccccccc}, 
      cell{2}{1} = {r=2}{},
      cell{4}{1} = {r=4}{},
      cell{8}{1} = {r=5}{},
      cell{14}{1} = {r=4}{},
      cell{18}{1} = {r=2}{},
      cell{21}{1} = {c=9}{},
      hline{1,21} = {-}{},
      hline{1,21} = {2}{-}{},
      hline{2} = {},
      hline{4} = {-}{},
      hline{4} = {2}{-}{},
      hline{8,13-14,18,20} = {dashed},
      cell{20}{3} = {font=\bfseries,Alto},
      cell{13}{4} = {font=\bfseries,Alto},
      cell{13}{5} = {font=\bfseries,Alto},
      cell{13}{6} = {font=\bfseries,Alto},
      cell{7}{7}  = {font=\bfseries,Alto},
      cell{13}{3} = {font=\itshape,Alto},
      cell{20}{4} = {font=\itshape,Alto},
      cell{20}{5} = {font=\itshape,Alto},
      cell{8}{6}  = {font=\itshape,Alto},
      cell{20}{7} = {font=\itshape,Alto},
    }
                                &           & \textbf{mAP} (\%) $\uparrow$ & \textbf{Coverage} $\downarrow$      & \textbf{Rankloss} (\%) $\downarrow$ & \textbf{OA} (\%) $\uparrow$       & \textbf{mF1} (\%) $\uparrow$     & \textbf{mprecision}  (\%) $\uparrow$   & \textbf{mrecall} (\%) $\uparrow$ \\
Ground-truth Training           & GT w/ BCE & 89.01 (0.32) & 5.834 (0.011) & 1.657 (0.061) & 94.29 (0.19) & 83.12 (0.66) & 88.87 (0.38) & 79.58 (1.28) \\
                                & IUN       & 82.24 (0.61) & 6.622 (0.031)  & 3.408 (0.026)  & 90.88 (0.19) & 72.65 (0.33) & 93.13 (0.86) & 62.87 (0.49) \\
Baseline Methods                & AN        & 64.89 (1.28) & 8.557 (0.164)  & 11.360 (0.351) & 71.60 (0.23) & 26.83 (0.18) & 86.05 (2.18) & 16.71 (0.13) \\
                                & AN-LS     & 64.40 (0.73) & 11.347 (0.231) & 21.664 (1.657) & 70.65 (0.28) & 19.97 (0.63) & 79.98 (5.71) & 12.14 (0.47) \\
                                & WAN       & 68.12 (0.46) & 7.620 (0.173)  & 8.055 (0.588)  & 78.06 (0.16) & 49.53 (0.42) & 84.19 (1.05) & 36.45 (0.27) \\
                                & EPR\textsuperscript{a}       & 68.38 (1.84) & 7.395 (0.079)  & 9.086 (0.412)  & 84.76 (0.57) & 69.82 (1.10) & 62.30 (0.85) & 81.11 (1.68) \\
General Label Noise Methods     & ELR       & 69.13 (0.73) & 7.469 (0.063)  & 5.914 (0.152)  & 87.99 (0.12) & 59.04 (0.84) & 74.80 (1.00) & 55.91 (0.30) \\
                                & LL-Cp     & 64.42 (1.06) & 7.792 (0.141)  & 7.931 (0.471)  & 78.64 (0.58) & 45.47 (1.25) & 76.83 (3.94) & 36.25 (0.78) \\
                                & LL-Ct     & 66.96 (1.60) & 8.574 (0.243)  & 7.647 (0.322)  & 85.42 (0.65) & 55.41 (0.91) & 78.43 (4.80) & 47.92 (0.75) \\
                                & LL-R      & 68.85 (0.44) & 8.526 (0.018)  & 8.616 (0.294)  & 80.39 (0.58) & 50.89 (2.19) & 84.92 (2.85) & 38.75 (2.31) \\
                                & BoostLU   & 68.00 (0.97) & 8.104 (0.127)  & 6.656 (0.519)  & 85.44 (0.97) & 54.48 (0.86) & 79.19 (1.90) & 47.14 (1.00) \\
PUL Methods                     & Dist-PU\textsuperscript{a}   & 70.85 (0.47) & 6.692 (0.133)  & 4.355 (0.265)  & 89.86 (0.09) & 60.48 (3.26) & 78.55 (6.93) & 57.04 (3.10) \\
SPML Methods                    & ROLE      & 56.35 (0.77) & 13.581 (0.041) & 19.800 (0.282) & 76.10 (0.38) & 43.26 (0.73) & 83.73 (3.32) & 30.84 (0.57) \\
                                & GR        & 62.14 (0.73) & 11.236 (0.090) & 21.537 (0.567) & 72.11 (0.09) & 28.32 (0.56) & 82.59 (4.41) & 18.18 (0.32) \\
                                & LAC       & 62.27 (0.92) & 7.902 (0.091)  & 9.674 (0.266)  & 72.57 (0.01) & 28.59 (1.81) & 82.17 (4.10) & 18.38 (1.22) \\
                                & MIME      & 61.55 (1.86) & 7.308 (0.175)  & 7.474 (0.756)  & 73.67 (0.26) & 31.55 (1.32) & 81.05 (3.42) & 20.52 (0.86) \\
\cy{RS-specific Methods}         & \cy{RCML} & 64.24 (1.41) & 7.522 (0.107)  & 8.338 (0.445)  & 80.10 (0.64) & 45.45 (0.47) & 75.62 (4.66) & 37.84 (0.23) \\
                                & \cy{LCR}  & 62.62 (0.47) & 8.269 (0.165)  & 10.453 (0.500) & 71.68 (0.13) & 23.48 (1.01) & 78.12 (2.50) & 15.06 (0.58) \\
Proposed                        & AdaGC     & 75.67 (0.26) & 6.862 (0.209)  & 5.636 (1.303)  & 83.39 (2.40) & 64.16 (1.57) & 68.89 (5.58) & 71.82 (6.80) \\
\textsuperscript{a} EPR and Dist-PU use additional a-priori information, which is assumed unknown in all other approaches.
    \end{tblr}}
\end{table*}

Table \ref{tab:ben-rand-results} presents the results on the \ac{reBEN} dataset under \textit{Random} \ac{SPML} noise. \ac{AdaGC} delivers the strongest overall performance, achieving the highest \ac{mAP}, \ac{OA}, and \ac{mF1}, together with the best coverage and rankloss scores. Although not optimal in mprecision or mrecall individually, it provides the most effective precision–recall trade-off, as reflected by its \ac{mF1}. General label-noise and \ac{PUL} methods perform well in this setting, often surpassing \ac{SPML}-specific approaches. \ac{WAN} remains particularly competitive among baselines. \cy{Within RS-specific methods, RCML offers noticeable improvements compared to baselines, whereas LCR contributes little due to its fixed class-correlation assumptions.} Compared with \ac{GT} training, \ac{AdaGC} closely approaches the full-label upper bound, especially in \ac{mF1}. Compared to \ac{IUN}, AdaGC attains slightly lower \ac{mAP} but higher \ac{mF1}. This highlights the importance of incorporating pseudo-labeling or label-correction mechanisms to recover missing labels and avoid confirmation bias.

\subsubsection{\ac{reBEN}-\textit{Dominant}}
\textit{Random} \ac{SPML} provides a simplified single-positive noise setting, whereas \textit{Dominant} \ac{SPML} is a more practical scenario driven by the visually dominant class in each image, resulting in systematic omission of non-dominant classes. This significantly increases task difficulty, as reflected in Table~\ref{tab:ben-domi-results}, where all methods exhibit lower absolute performance compared with Table~\ref{tab:ben-rand-results}. Despite this challenge, \ac{AdaGC} remains the strongest method, achieving the best \ac{mAP}, \ac{mF1}, coverage, and rankloss. In contrast, most competing approaches fail to surpass the \ac{WAN} and \ac{EPR} baselines. \ac{WAN} obtains the second-best \ac{mAP} and outperforms all \ac{SPML}-specific methods in \ac{mF1}. LL-based methods and Dist-PU offer additional gains over \ac{WAN} in \ac{mF1}, yet still fall short of \ac{EPR}. Thanks to its cardinality regularization, \ac{EPR} effectively mitigates the strong bias induced by \textit{Dominant} \ac{SPML}, yielding competitive \ac{mF1} despite its low \ac{mAP}. \cy{RCML remains relatively strong but continues to underperform the proposed \ac{AdaGC}.}

\subsubsection{AID-\textit{Random}}
\begin{figure}[t]
    \subfigure[\ac{reBEN}]{\includegraphics[width=0.48\columnwidth]{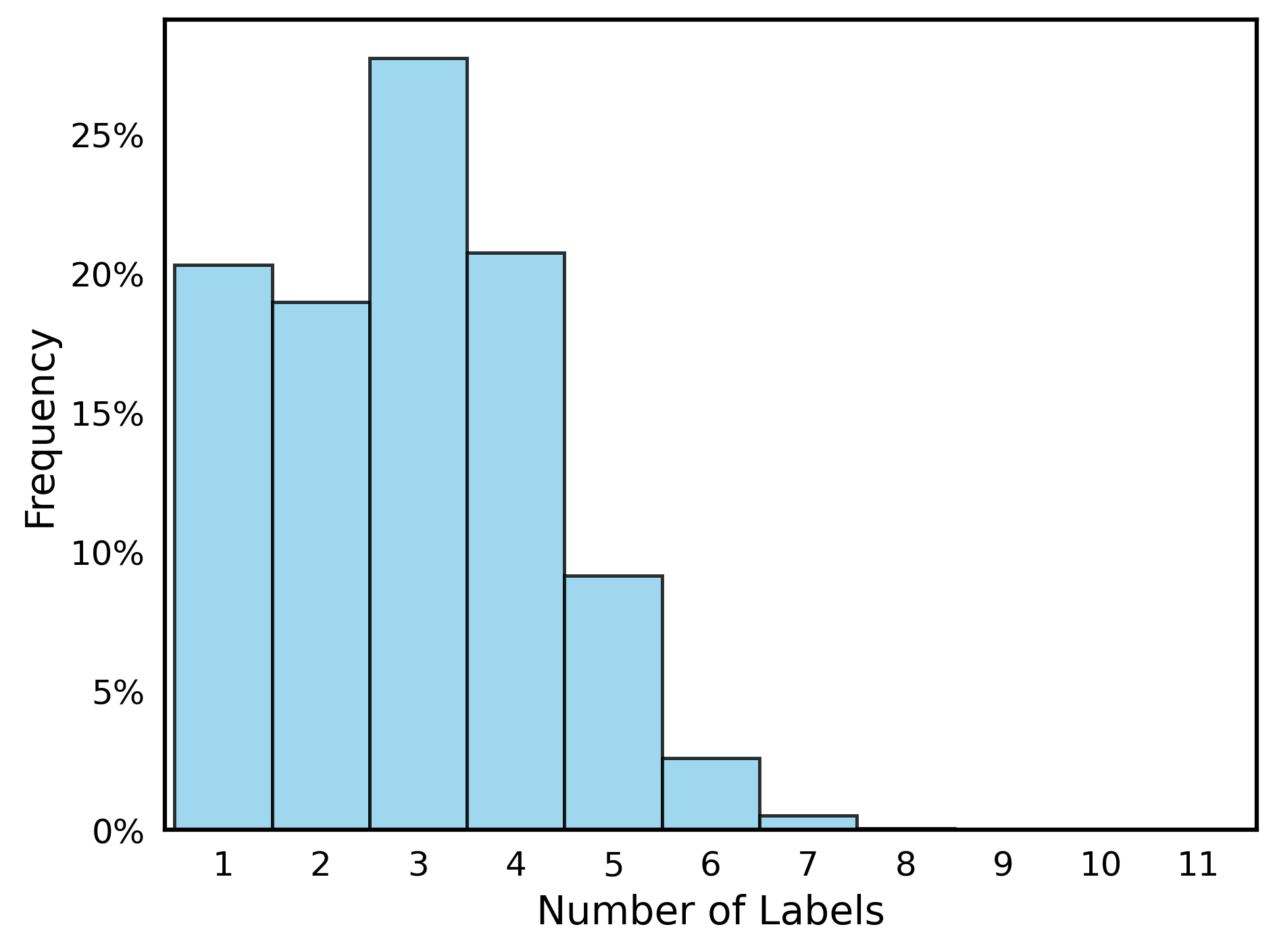}}
    \subfigure[AID]{\includegraphics[width=0.48\columnwidth]{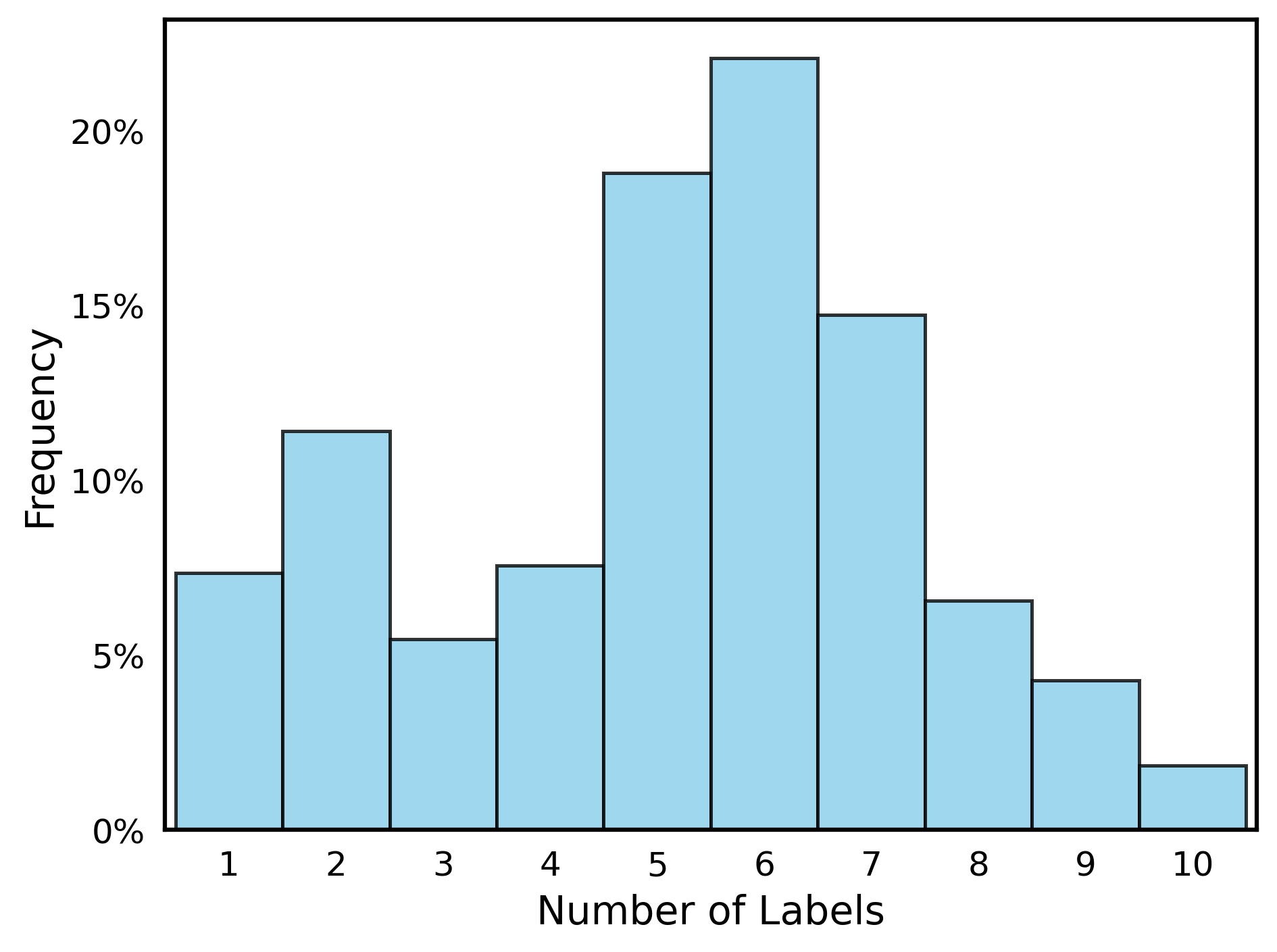}}
    \caption{Histograms of the number of \ac{GT} labels per image in the training sets.}\label{fig:labelhist}
\end{figure}
Table \ref{tab:aid-results} reports the results on the AID dataset under \textit{Random} \ac{SPML} noise. Owing to ImageNet pre-training and richer spatial information, this dataset yields higher absolute performance than \ac{reBEN}, despite larger flip rates $\beta_c$. \ac{AdaGC} shows the best performance across all metrics. It achieves the highest \ac{mAP} and the second-best \ac{mF1}, mrecall, coverage, and rankloss scores. Dist-PU also shows excellent results, matching \ac{AdaGC} in \ac{mF1} while achieving the best \ac{OA}, coverage, and rankloss. Meanwhile, \ac{EPR} reaches the highest \ac{mF1} overall. The strong outcomes of Dist-PU and \ac{EPR} are largely attributable to their use of prior information about the underlying \ac{GT} label distributions, which is not available to \ac{AdaGC}. Dist-PU assumes knowledge of class-prior distributions and aligns predicted label frequencies accordingly, whereas \ac{EPR} relies on the expected number of positive labels per image. These priors are particularly beneficial for AID, which contains more labels per image on average (Fig.~\ref{fig:labelhist}). As a result, \ac{EPR} achieves the highest mrecall, although this comes at the cost of the lowest mprecision, reflecting its tendency to overestimate positives. \cy{Still, RCML achieves some comparable results to other considered methods, whereas the improvement brought by LCR remains limited.}

\subsubsection{\cy{AID-\textit{Manual}}}
\begin{table*}
    \centering
    \caption{\cy{Test Performance Comparison of Different Methods on the AID-Manual Dataset. The Best Average Metric Values Are Reported in Bold, Second Bests Are in Italic. The Related Standard Deviations Are Reported in Brackets}}
    \label{tab:aid-results-manual}
    \resizebox{\textwidth}{!}{%
    \begin{tblr}{
      colspec = {llccccccc}, 
      cell{2}{1} = {r=2}{},
      cell{4}{1} = {r=4}{},
      cell{8}{1} = {r=5}{},
      cell{14}{1} = {r=4}{},
      cell{18}{1} = {r=2}{},
      cell{21}{1} = {c=9}{},
      hline{1,21} = {-}{},
      hline{1,21} = {2}{-}{},
      hline{2} = {},
      hline{4} = {-}{},
      hline{4} = {2}{-}{},
      hline{8,13-14,18,20} = {dashed},
      cell{20}{3} = {font=\bfseries,Alto},
      cell{13}{4} = {font=\bfseries,Alto},
      cell{13}{5} = {font=\bfseries,Alto},
      cell{13}{6} = {font=\bfseries,Alto},
      cell{20}{7} = {font=\bfseries,Alto},
      cell{8}{3}  = {font=\itshape,Alto},
      cell{20}{4} = {font=\itshape,Alto},
      cell{20}{5} = {font=\itshape,Alto},
      cell{8}{6}  = {font=\itshape,Alto},
      cell{7}{7}  = {font=\itshape,Alto},
    }
                                &           & \textbf{mAP} (\%) $\uparrow$ & \textbf{Coverage} $\downarrow$      & \textbf{Rankloss} (\%) $\downarrow$ & \textbf{OA} (\%) $\uparrow$       & \textbf{mF1} (\%) $\uparrow$     & \textbf{mprecision}  (\%) $\uparrow$   & \textbf{mrecall} (\%) $\uparrow$ \\
Ground-truth Training           & GT w/ BCE & 89.01 (0.32) & 5.834 (0.011) & 1.657 (0.061) & 94.29 (0.19) & 83.12 (0.66) & 88.87 (0.38) & 79.58 (1.28) \\
                                & IUN     & 82.71   (0.60) & 6.638 (0.085)  & 3.068 (0.191)  & 91.47 (0.16) & 74.83 (0.42) & 94.18 (3.95) & 65.53 (0.89) \\
Baseline Methods                & AN      & 67.66 (0.73)   & 8.589 (0.140)  & 11.276 (0.283) & 71.30 (0.08) & 29.01 (0.59) & 83.94 (1.69) & 18.83 (0.56) \\
                                & AN-LS   & 66.75 (0.62)   & 11.556 (0.061) & 23.300 (0.640) & 71.34 (0.17) & 28.41 (1.81) & 84.61 (1.94) & 18.32 (1.19) \\
                                & WAN     & 70.28 (0.50)   & 7.721 (0.103)  & 8.364 (0.276)  & 75.35 (0.29) & 45.28 (1.77) & 88.30 (4.71) & 32.21 (1.68) \\
                                & EPR\textsuperscript{a}     & 71.92 (1.36)   & 7.251 (0.118)  & 8.766 (0.358)  & 84.89 (0.50) & 71.30 (1.06) & 63.86 (1.49) & 83.84 (0.56) \\
General Label Noise Methods     & ELR     & 76.99 (0.51)   & 7.278 (0.107)  & 5.353 (0.232) & 88.26 (0.31) & 69.09 (0.63) & 77.43 (0.47) & 68.55 (1.33) \\
                                & LL-Cp   & 69.65 (0.54)   & 7.854 (0.073)  & 8.982 (0.470)  & 70.87 (0.07) & 23.14 (0.23) & 80.08 (3.81) & 14.94 (0.28) \\
                                & LL-Ct   & 72.46 (0.11)   & 7.564 (0.104)  & 7.788 (0.525)  & 76.04 (0.12) & 42.70 (0.69) & 85.52 (2.13) & 30.93 (0.24) \\
                                & LL-R    & 73.89 (0.29)   & 7.743 (0.007)  & 6.560 (0.094)  & 83.95 (0.05) & 63.51 (1.55) & 80.53 (1.29) & 54.60 (1.90) \\
                                & BoostLU & 71.35 (1.77)   & 8.441 (0.061)  & 8.976 (0.247)  & 75.61 (0.06) & 42.58 (0.40) & 83.31 (6.37) & 31.62 (0.56) \\
PUL Methods                     & Dist-PU\textsuperscript{a} & 74.90 (1.88)   & 6.574 (0.021)  & 3.652 (0.085)  & 90.23 (0.34) & 66.77 (1.16) & 80.98 (0.23) & 62.70 (1.32) \\
SPML Methods                    & ROLE    & 64.36 (1.99)   & 13.253 (0.101) & 17.429 (0.973) & 76.41 (0.64) & 49.11 (2.03) & 86.08 (4.40) & 36.30 (2.01) \\
                                & GR      & 67.18 (0.65)   & 11.042 (0.191) & 21.481 (0.353) & 71.63 (0.14) & 29.88 (1.34) & 86.64 (4.63) & 19.93 (1.17) \\
                                & LAC     & 65.80 (0.81)   & 7.878 (0.066)  & 9.395 (0.259)  & 72.64 (0.16) & 31.72 (1.58) & 80.81 (4.74) & 21.03 (1.04) \\
                                & MIME    & 65.39 (1.07)   & 7.240 (0.147)  & 6.333 (0.424)  & 73.10 (0.14) & 32.62 (0.89) & 84.73 (6.70) & 21.70 (0.65) \\
\cy{RS-specific Methods}         & \cy{RCML} & 72.87 (0.57) & 7.453 (0.019) & 7.393 (0.266) & 80.46 (1.44) & 52.99 (3.67) & 85.83 (2.34) & 43.08 (5.79) \\
                                & \cy{LCR}  & 66.61 (0,96) & 8.322 (0.078) & 10.669 (0.472)  & 71.79 (0.22) & 31.83 (1.46) & 85.52 (1.57) & 21.03 (1.02) \\
Proposed                        & AdaGC     & 79.07 (0.91) & 6.673 (0.103) & 4.620 (0.558) & 82.66 (1.01)  & 71.95 (0.94) & 68.25 (2.35) & 82.61 (1.28) \\
\textsuperscript{a} EPR and Dist-PU use additional a-priori information, which is assumed unknown in all other approaches.
    \end{tblr}}
\end{table*}
\cy{Table \ref{tab:aid-results-manual} reports the results on the AID dataset using \textit{Manual} \ac{SPML} labels to further evaluate AdaGC’s practical effectiveness. Benefiting from human-induced bias in single-positive labels, most methods achieve higher performance compared with AID-\textit{Random}. In this setting, AdaGC remains highly effective, getting the highest \ac{mAP} and \ac{mF1}, and ranking among the top methods across nearly all metrics. Dist-PU again achieves the best \ac{OA}, coverage, and rankloss by leveraging extra prior information. RS-specific methods show similar trends as before. RCML provides moderate improvements over other SPML or label-noise baselines, while LCR contributes little due to the fixed class-correlation assumptions. Overall, the AID-Manual results confirm that higher-quality single-positive labels enhance learning, and AdaGC consistently delivers robust performance without relying on additional priors.}

\subsection{Ablation Experiments} \label{sec:experiments:ablation}
We conduct ablation experiments on \ac{reBEN} under both types of label noise to evaluate the effectiveness of individual components in \ac{AdaGC}. Without making specific claims, we report the accuracies of the teacher models below.

\subsubsection{\cy{Impact of Early Learning Detection}}
\begin{figure*}[ht]
    \subfigure[\textit{Random}: mAP]{\includegraphics[width=0.49\columnwidth]{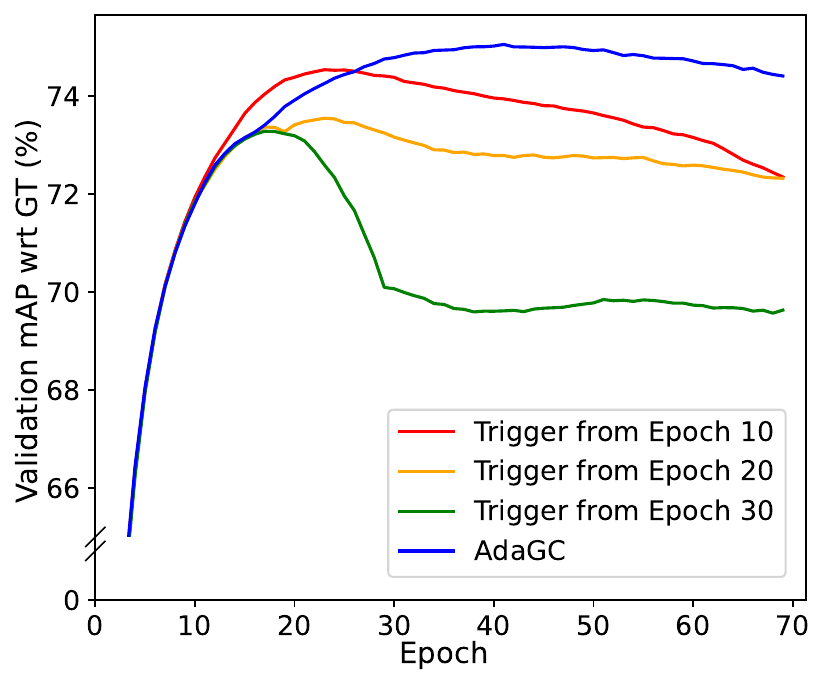}}
    \subfigure[\textit{Random}: mF1]{\includegraphics[width=0.49\columnwidth]{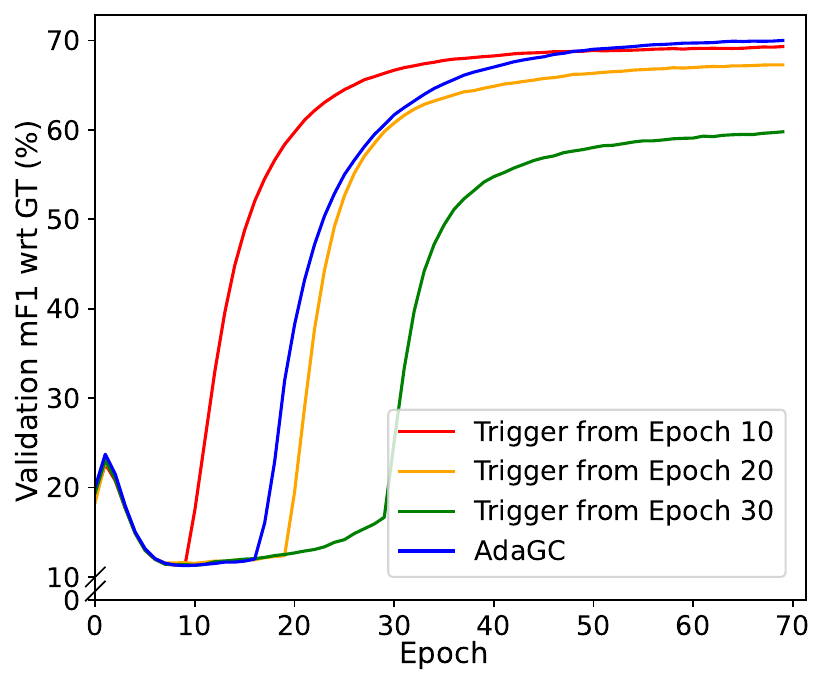}}
    \subfigure[\textit{Dominant}: mAP]{\includegraphics[width=0.49\columnwidth]{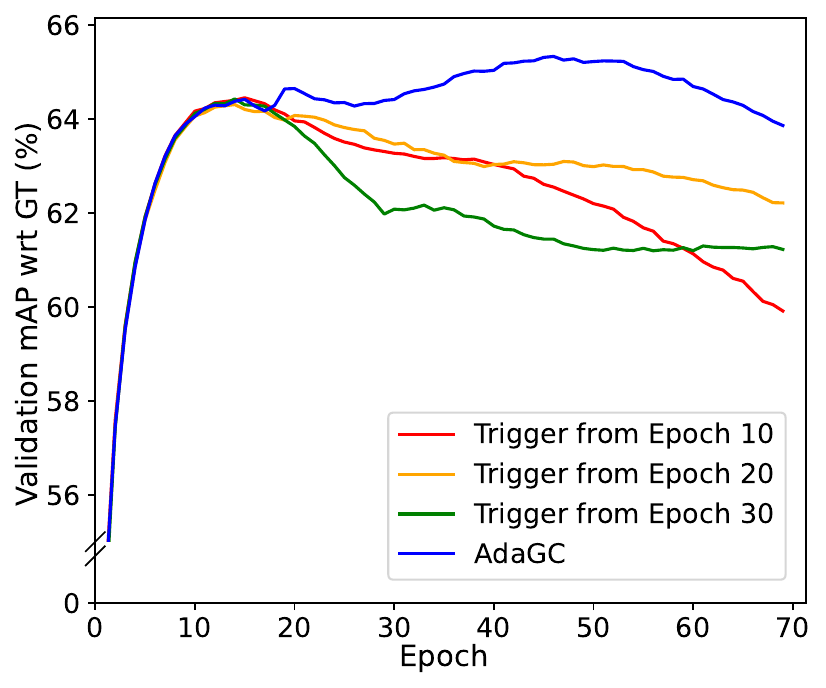}}
    \subfigure[\textit{Dominant}: mF1]{\includegraphics[width=0.49\columnwidth]{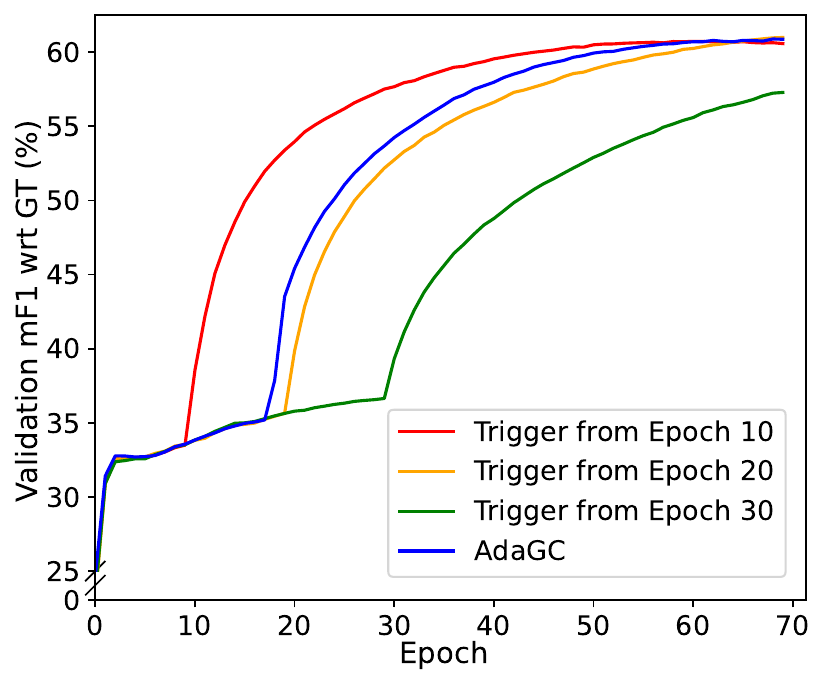}}
    \caption{\cy{Validation accuracies (mAP and mF1) over training epochs obtained with predefined warm-up lengths (10, 20, 30) and our proposed early learning detection strategy (AdaGC).}}
    \label{fig:exp:earlylearn}
\end{figure*}
\cy{We evaluate the effectiveness of early learning detection for adaptively triggering \ac{GC} by comparing AdaGC with variants that use manually predefined trigger points. Fig.~\ref{fig:exp:earlylearn} plots their validation accuracies (wrt GT labels) over training time. The proposed adaptive detection strategy successfully activates \ac{GC} at an appropriate point, resulting in the most significant performance improvements. This demonstrates that the timing of GC activation is essential for model convergence stability. In contrast, triggering \ac{GC} too early or too late leads to suboptimal results. Notably, activating \ac{GC} too late leads to even worse performance, highlighting the detrimental effect of overfitting to label noise. Furthermore, we report their final test accuracies in Table~\ref{tab:abla:earlylearning}. AdaGC achieves competitive or comparable \ac{mAP} and \ac{mF1} scores compared to those obtained using fixed warm-up lengths, indicating the robustness of the proposed early learning detection strategy.}

\begin{table}[t]
\centering
\setlength{\extrarowheight}{0.2mm}
\setlength\tabcolsep{6.pt}
\caption{Accuracies (\%) on the Test Set Obtained With Different GC Trigger Settings After 70 Epochs}
\label{tab:abla:earlylearning}
\begin{tabular}{c|cc|cc}
\hline\hline
                        & \multicolumn{2}{c|}{\textit{Random}} & \multicolumn{2}{c}{\textit{Dominant}} \\
                        \cline{2-5}
                        & \textbf{mAP} $\uparrow$          & \textbf{mF1} $\uparrow$         & \textbf{mAP} $\uparrow$           & \textbf{mF1} $\uparrow$           \\
\hline
AdaGC (after $\sim$17 epochs)               & 68.18        & 64.65       & 60.22         & 58.78        \\
\hdashline
Trigger after 10 epochs & 65.78        & 64.06       & 57.16         & 58.58        \\
Trigger after 20 epochs & 65.87        & 62.39       & 59.35         & 59.28        \\
Trigger after 30 epochs & 63.18        & 54.77       & 58.62         & 55.71    \\
\hline\hline
\end{tabular}
\end{table}

\begin{figure}
    \subfigure[\textit{Random}]{\includegraphics[width=0.49\columnwidth]{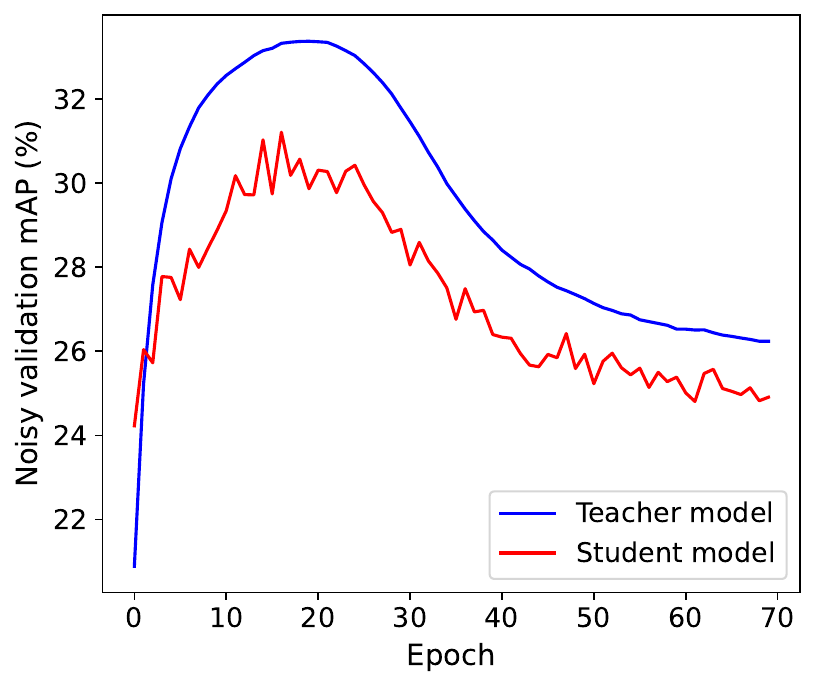}}
    \subfigure[\textit{Dominant}]{\includegraphics[width=0.49\columnwidth]{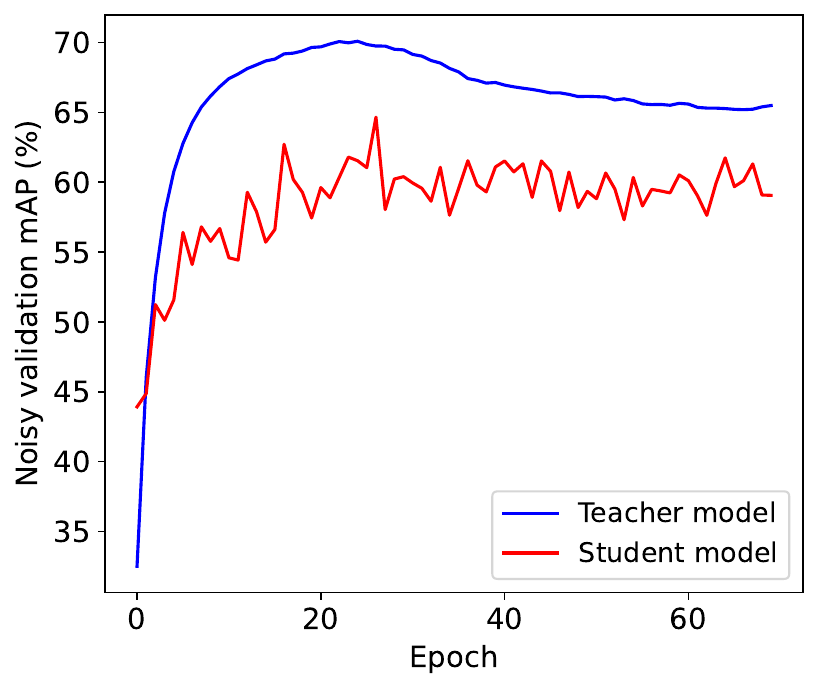}}
    \caption{Noisy validation \ac{mAP} (with respect to noisy labels) versus training time (epoch) obtained by AN on the \ac{reBEN} dataset.}
    \label{fig:exp:abla:AN}
\end{figure}

\cy{We demonstrate the necessity of using the teacher model for early learning detection by comparing the smoothness of noisy validation accuracies from the student and teacher models in Fig.~\ref{fig:exp:abla:AN}. Although both models exhibit similar overall trends, the teacher model produces noticeably smoother accuracy curves. This stability makes it more suitable for reliable early learning detection. In contrast, detections based on the student model are more prone to local maxima, resulting in unstable and suboptimal outcomes.}

\subsubsection{\cy{Effect of the Dual-EMA Mechanism}}

\begin{figure*}[ht]
    \subfigure[\textit{Random}: pseudo-label mAP]{\includegraphics[width=0.49\columnwidth]{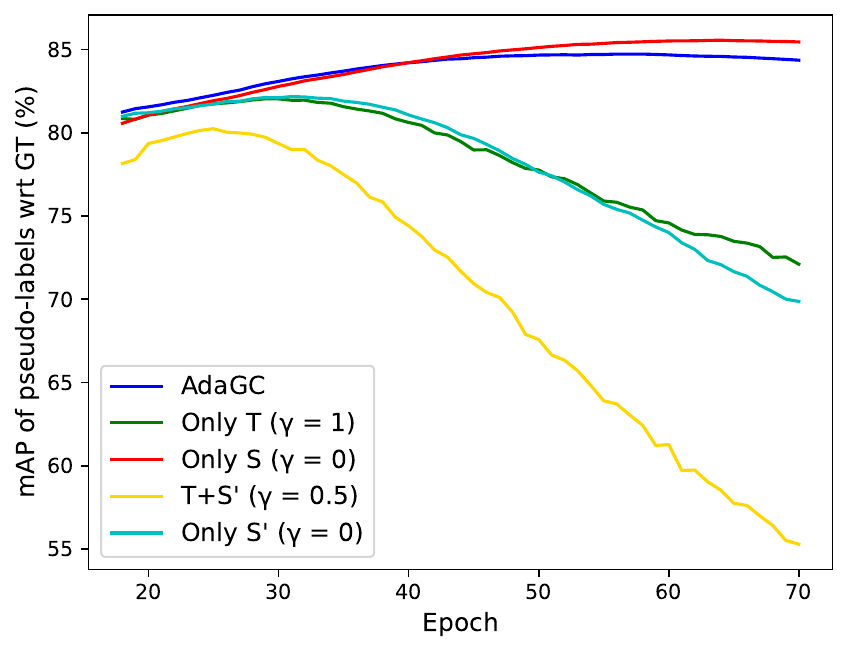}}
    \subfigure[\textit{Random}: pseudo-label mF1]{\includegraphics[width=0.49\columnwidth]{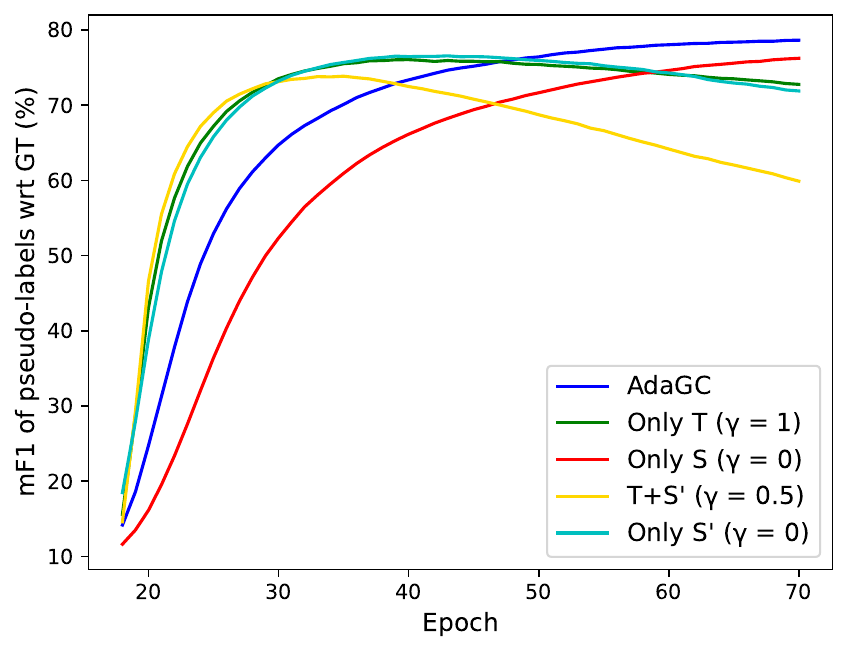}}
    \subfigure[\textit{Random}: validation mAP]{\includegraphics[width=0.49\columnwidth]{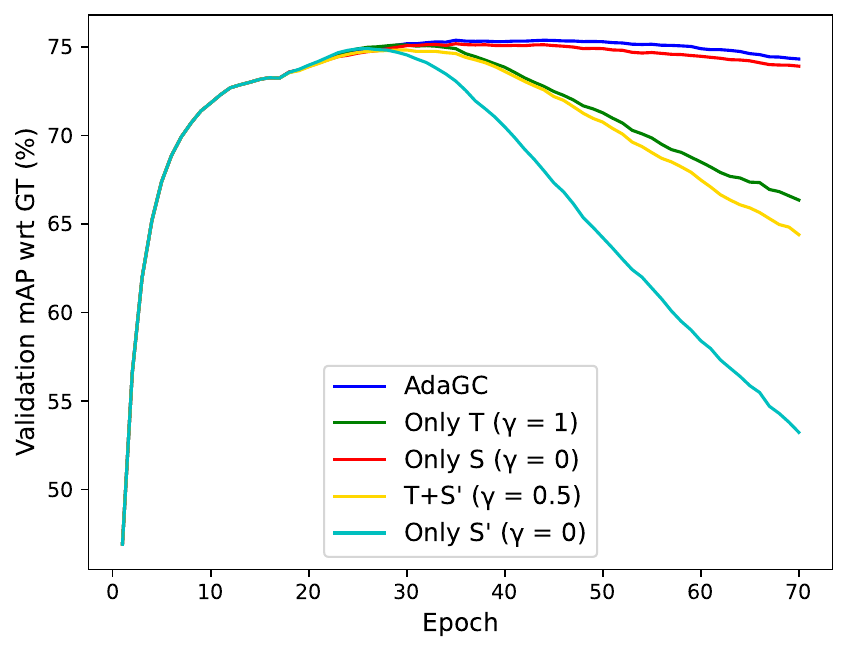}}
    \subfigure[\textit{Random}: validation mF1]{\includegraphics[width=0.49\columnwidth]{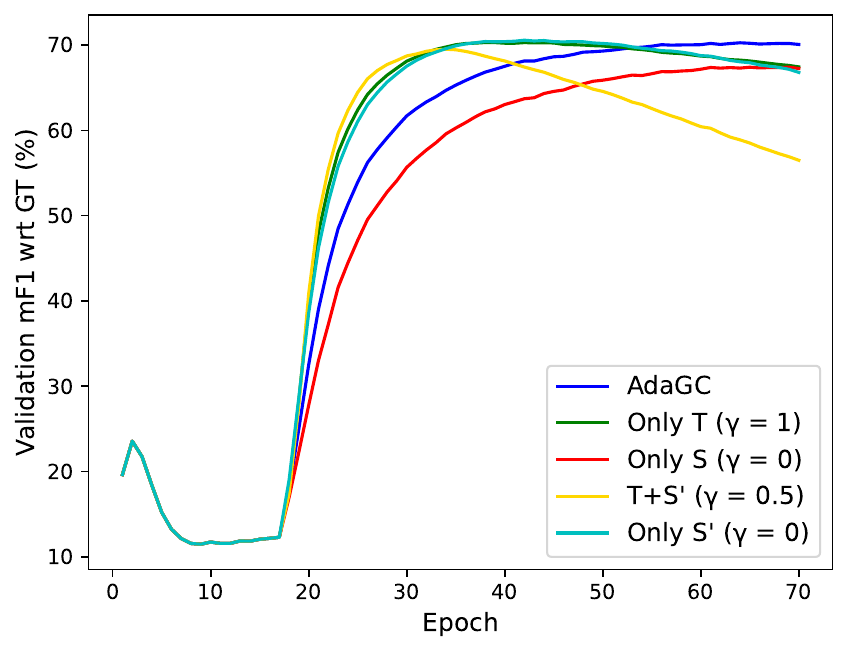}}\\
    \subfigure[\textit{Dominant}: pseudo-label mAP]{\includegraphics[width=0.49\columnwidth]{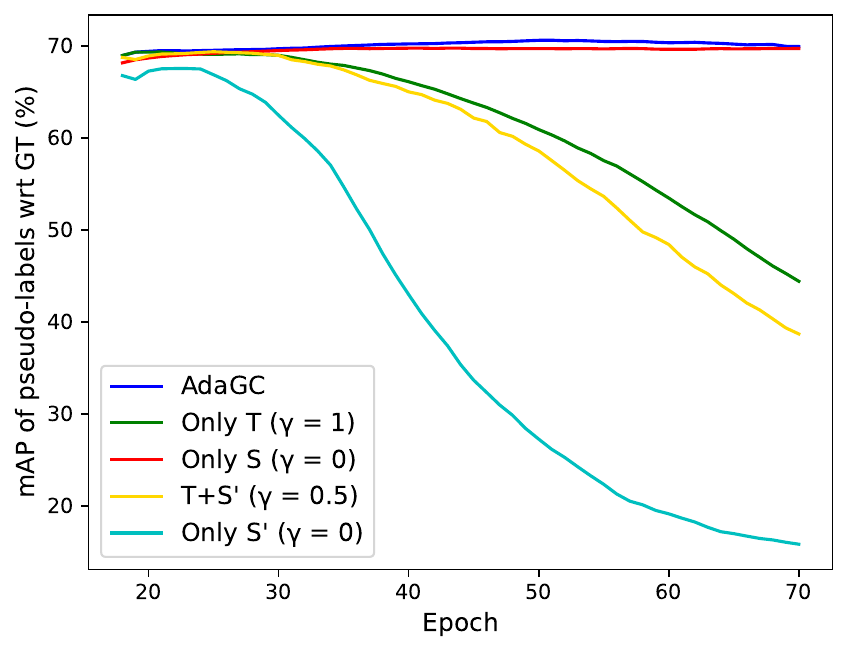}}
    \subfigure[\textit{Dominant}: pseudo-label mF1]{\includegraphics[width=0.49\columnwidth]{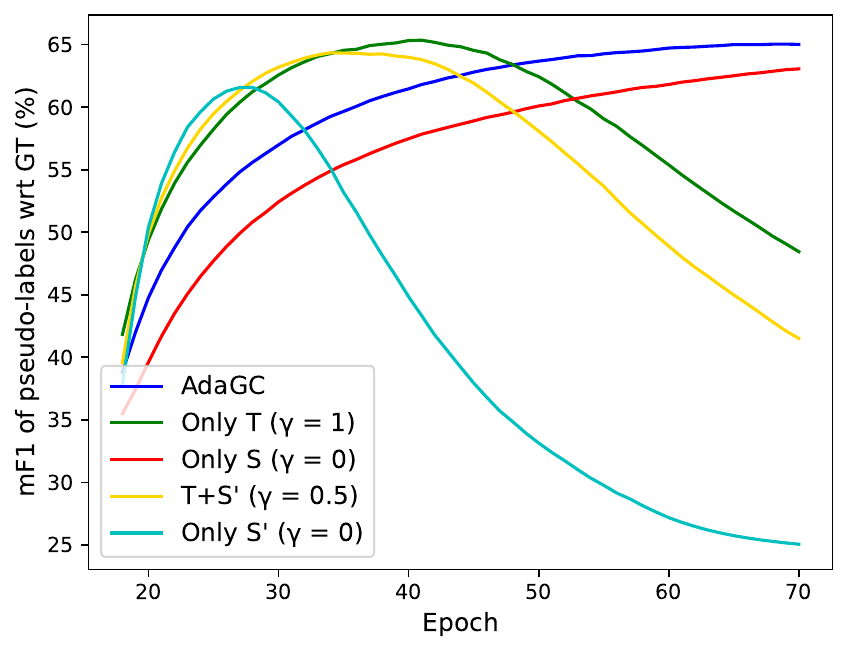}}
    \subfigure[\textit{Dominant}: validation mAP]{\includegraphics[width=0.49\columnwidth]{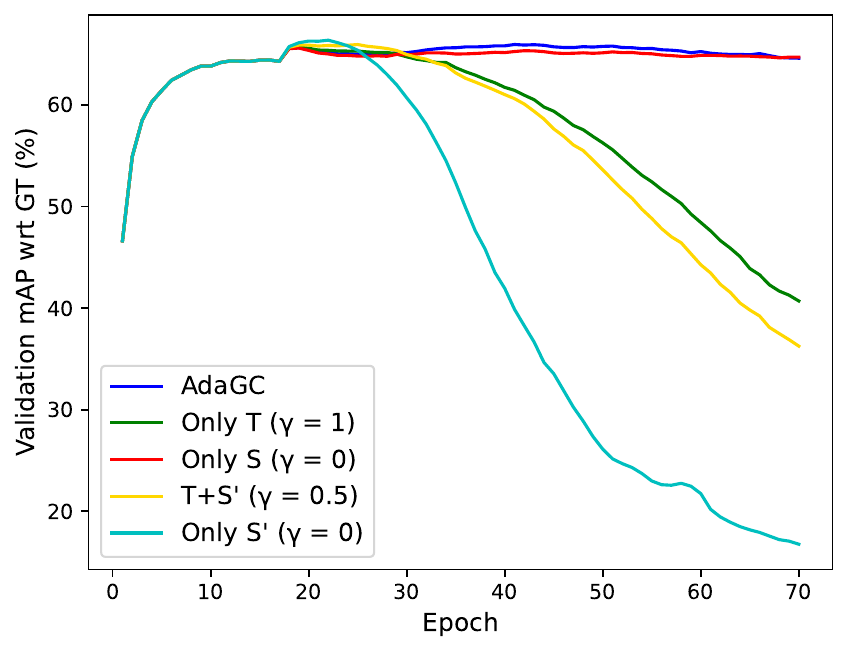}}
    \subfigure[\textit{Dominant}: validation mF1]{\includegraphics[width=0.49\columnwidth]{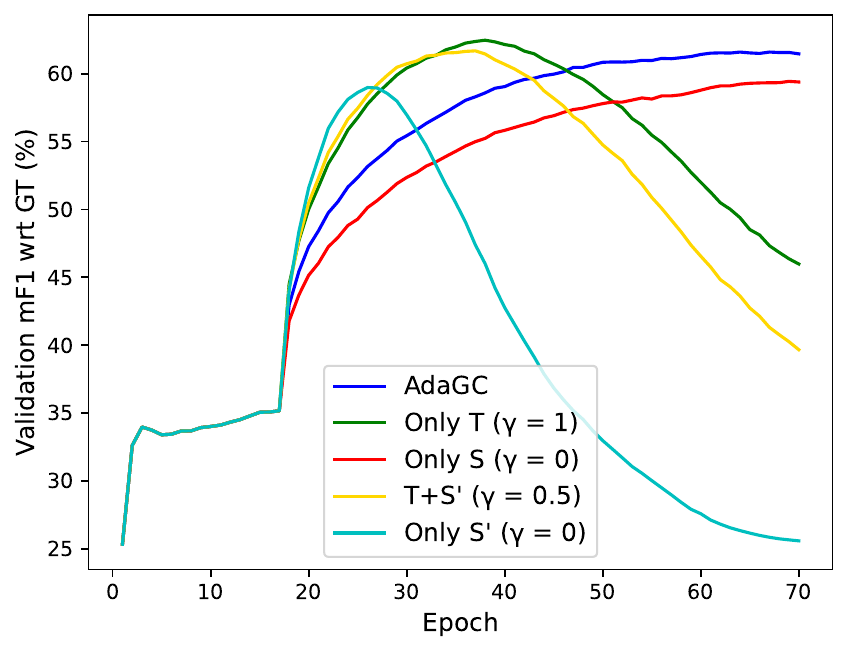}}
    \caption{\cy{Pseudo-label quality (a), (b), (e), (f) and AdaGC validation accuracies (mAP and mF1) (c), (d), (g), (h) over training epochs using different pseudo-label generation strategies, where T, S, and S$'$ denote teacher model predictions, EMA-smoothed student model predictions, and student model predictions without EMA, respectively. (a)-(d) Random single-positive labels case. (e)-(h) Dominant single-positive labels case.}}
    \label{fig:exp:pseudo}
\end{figure*}

\cy{We investigate the effect of the proposed dual-EMA mechanism by comparing it with single-EMA and no-EMA settings, as shown in Table~\ref{tab:abla:pseudo}. The dual-EMA strategy consistently achieves the most robust performance by striking a balance between stability and adaptability. In contrast, the single-EMA and no-EMA variants exhibit suboptimal results, particularly when the EMA smoothing on student predictions is removed. In both \textit{Random} and \textit{Dominant} noise settings, the EMA-smoothed student predictions are more stable than the teacher-only ones, owing to the teacher’s slower update rate. Nevertheless, without EMA smoothing (no-EMA), the student’s raw predictions tend to amplify errors, resulting in additional performance degradation. This trend is further confirmed in Fig.~\ref{fig:exp:pseudo}, where pseudo-label quality is shown to have a strong impact on AdaGC’s performance. As the pseudo-label quality deteriorates, performance drops markedly, especially for the teacher-only and no-EMA cases. The dual-EMA mechanism mitigates these issues by smoothing student predictions and integrating teacher outputs, thereby generating more reliable pseudo labels. Considering the varying update dynamics of the student and teacher models during training, a more adaptive combination strategy beyond the current half–half scheme is expected to enhance robustness, which we leave for future exploration.}

\begin{table}[t]
\centering
\setlength{\extrarowheight}{0.5mm}
\caption{
\cy{Test set accuracies (\%) obtained using different pseudo-label generation strategies, where T, S, and S$'$ denote teacher model predictions, EMA-smoothed student model predictions, and student model predictions without EMA, respectively}}
\label{tab:abla:pseudo}
\begin{tabular}{c|c|cc|cc}
\hline\hline
\multicolumn{2}{c|}{\multirow{2}{*}{EMA Settings}}  & \multicolumn{2}{c|}{\textit{Random}} & \multicolumn{2}{c}{\textit{Dominant}} \\
\cline{3-6}
\multicolumn{2}{c|}{} & \textbf{mAP} $\uparrow$          & \textbf{mF1} $\uparrow$         & \textbf{mAP} $\uparrow$           & \textbf{mF1} $\uparrow$          \\
\hline
dual & AdaGC  & 68.18        & 64.65       & 60.22         & 58.78        \\
\hdashline
\multirow{3}{*}{single} & Only T ($\gamma=1$)       & 61.74        & 63.49       & 42.01         & 46.40        \\
& Only S ($\gamma=0$)       & 67.93        & 61.52       & 60.11         & 57.42       \\
& T+S$'$ ($\gamma=0.5$)       & 53.78        & 61.94       & 27.50         & 35.71       \\
\hdashline
no & Only S$'$ ($\gamma=0$)       & 49.30        & 52.89       & 16.41         & 25.24       \\
\hline\hline
\end{tabular}
\end{table}

\subsubsection{\cy{Role of Mixup}}

\cy{We then investigate the effect of Mixup under both \textit{Random} and \textit{Dominant} noise settings. As shown in Table~\ref{tab:abla:mixup}, removing Mixup consistently reduces performance, particularly in the \textit{Random} case, with about a 2\% drop in \ac{mAP} and \ac{mF1}. This indicates that Mixup enhances generalization by enforcing smoother decision boundaries. It can complement the GC mechanism in improving the robustness of AdaGC.}

\begin{table}[t]
\centering
\setlength{\extrarowheight}{0.2mm}
\setlength\tabcolsep{10pt}
\caption{\cy{Test accuracies (\%) Obtained With and Without Mixup}}
\label{tab:abla:mixup}
\begin{tabular}{c|cc|cc}
\hline\hline
      & \multicolumn{2}{c|}{\textit{Random}} & \multicolumn{2}{c}{\textit{Dominant}} \\
      \cline{2-5}
      & \textbf{mAP} $\uparrow$          & \textbf{mF1} $\uparrow$         & \textbf{mAP} $\uparrow$           & \textbf{mF1} $\uparrow$      \\
\hline
AdaGC & 68.18        & 64.65       & 60.22         & 58.78        \\
\hdashline
w/o Mixup & 66.00        & 62.08       & 59.05         & 59.19       \\
\hline\hline
\end{tabular}
\end{table}

\section{Conclusion} \label{sec:con}

In this work, we have presented \ac{AdaGC}, a novel and generalizable framework for \ac{SPML} tailored to \ac{RS} imagery, which addresses overfitting to false negatives via a gradient calibration mechanism. To ensure its operational effectiveness, we introduce a simple yet theoretically grounded indicator to adaptively activate GC when it detects the end of early learning during an initial warm-up stage. In addition, a dual-EMA module is introduced to produce robust pseudo-labels that support the GC process, along with Mixup to further boost the robustness of training.

Extensive experiments on two benchmark RS datasets under two single-positive realistic label noise simulation settings, \textit{Random} and \textit{Dominant}, demonstrate the effectiveness and robustness of \ac{AdaGC}. \ac{AdaGC} consistently outperforms existing SPML baselines across different scenarios. Our work fills a critical gap in the literature, where SPML remains largely unexplored in the RS domain. While \ac{AdaGC} performs well in most settings, it shows relative limitations under the dominant single-positive label noise condition. As this scenario is common in many real-world annotation practices, future work will focus on enhancing \ac{AdaGC}'s performance under such conditions, further advancing scalable and reliable multi-label learning in \ac{RS}.

\bibliographystyle{IEEEtran}
\bibliography{references}

\end{document}